\documentclass{article}
\usepackage{microtype}
\usepackage{graphicx}
\usepackage{subfigure}
\usepackage{booktabs}
\usepackage{xfrac}
\usepackage{physics}
\usepackage{amsmath,amssymb,amsthm,mathrsfs, bm}
\usepackage{enumitem}
\usepackage{comment}
\usepackage{dsfont}

\newcommand{\iris}[1]{#1}

\usepackage{wrapfig}
\usepackage[utf8]{inputenc} 
\usepackage[T1]{fontenc}    
\usepackage{hyperref}       
\usepackage{url}            
\usepackage{booktabs}       
\usepackage{amsmath, amssymb, amsthm, amsfonts}       
\usepackage{nicefrac, xfrac}       
\usepackage{microtype}      
\usepackage{bm}
\usepackage{enumitem}
\usepackage{graphicx}
\usepackage{algorithm}
\usepackage{algorithmic}
\usepackage{geometry}
\usepackage{authblk,textcomp}
\usepackage{mathtools}
\usepackage{commath}
\usepackage{libertine}
\usepackage{mathrsfs}
\usepackage{enumitem}

\usepackage{amsmath}
\usepackage{amssymb}
\usepackage{mathtools}
\usepackage{amsthm}
\usepackage{bbm}

\hypersetup{pdfauthor={},pdftitle={},%
            colorlinks, linktocpage=true, pdfstartpage=1, pdfstartview=FitV,%
    breaklinks=true, pdfpagemode=UseNone, pageanchor=true, pdfpagemode=UseOutlines,%
    plainpages=false, bookmarksnumbered, bookmarksopen=true, bookmarksopenlevel=1,%
    hypertexnames=true, pdfhighlight=/O,%
    urlcolor=orange, linkcolor=blue, citecolor=blue
        }

\usepackage[dvipsnames]{xcolor}

\usepackage[numbers]{natbib}
\usepackage{amsmath}
\usepackage{multicol}   
\usepackage{enumitem}   
\usepackage[export]{adjustbox} 
\usepackage[utf8]{inputenc} 
\usepackage[T1]{fontenc}    
\usepackage{hyperref}       
\usepackage{url}            
\usepackage{booktabs}       
\usepackage{amsfonts}       
\usepackage{nicefrac}       
\usepackage{microtype}      
\usepackage[dvipsnames]{xcolor}         
\usepackage{graphicx}
\usepackage{wrapfig}

\title{Dataset Distillation for Memorized Data:\texorpdfstring{\\}{} Soft Labels can Leak Held-Out Teacher Knowledge}

\author[1]{Freya Behrens}
\author[1]{Lenka Zdeborov\'a}

\affil[1]{\small 
Statistical Physics of Computation Laboratory, École polytechnique fédérale de Lausanne (EPFL), Switzerland}
\affil[ ]{\texttt{\scriptsize\{freya.behrens,lenka.zdeborova\}@epfl.ch}}

\newcommand{\Dteacher}{\mathcal{D}^{\text{T}}_{\star}}
\newcommand{\Dstudenttrain}{\mathcal{D}^{\text{S}}_{\text{train}}}
\newcommand{\Dstudenttest}{\mathcal{D}^{\text{S}}_{\text{test}}}
\newcommand{\Dval}{\mathcal{D}_{\text{val}}}

\newcommand{\accTestStudent}{\mathrm{acc}^{\mathrm{S}}_{\mathrm{test}}}
\newcommand{\accTrainStudent}{\mathrm{acc}^{\mathrm{S}}_{\mathrm{train}}}
\newcommand{\accValStudent}{\mathrm{acc}^{\mathrm{S}}_{\mathrm{val}}}
\newcommand{\accValTeacher}{\mathrm{acc}^{\mathrm{T}}_{\mathrm{val}}}
\newcommand{\accTrainTeacher}{\mathrm{acc}_{\star}^{\mathrm{T}}}
\newcommand{\accZeroStudent}{\mathrm{acc}^{\mathrm{S}}_{\mathrm{c=1}}}
\newcommand{\alphaTlabel}{\alpha^{\text{T}}_{\text{label}}}
\newcommand{\alphaSid}{\alpha^{\text{S}}_{\text{id}}}
\newcommand{\alphaSlabel}{\alpha^{\text{S}}_{\text{label}}}
\newcommand{\alphaSlabelshuffle}{\alpha^{\text{S-shuffle}}_{\text{label}}}

\begin{document}

\date{}
\maketitle

\begin{abstract}
Dataset distillation aims to compress training data into fewer examples via a teacher, from which a student can learn effectively. While its success is often attributed to structure in the data, modern neural networks also memorize specific facts, but if and how such memorized information is can transferred in distillation settings remains less understood.
In this work, we show that students trained on soft labels from teachers can achieve non-trivial accuracy on held-out memorized data they never directly observed.
This effect persists on structured data when the teacher has not generalized.
To analyze it in isolation, we consider finite random i.i.d. datasets where generalization is a priori impossible and a successful teacher fit implies pure memorization.
Still, students can learn non-trivial information about the held-out data, in some cases up to perfect accuracy. 
In those settings, enough soft labels are available to recover the teacher functionally -- the student matches the teacher's predictions on all possible inputs, including the held-out memorized data.
We show that these phenomena strongly depend on the temperature with which the logits are smoothed, but persist across varying network capacities, architectures and dataset compositions.
\end{abstract}

\vspace{3em}
\section{Introduction}

With the advent of foundation models, it has become of great interest to exploit and transfer their capabilities to other models, e.g. via knowledge or dataset distillation.
The goal of knowledge distillation is to train a smaller student on a small amount of data derived from a teacher~\cite{xuSurveyKnowledgeDistillation2024}; dataset distillation focuses on finding a minimal training set that achieves high performance, possibly modifying the training set via information derived from the teacher~\citep{cazenavetteGeneralizingDatasetDistillation2023,yuDatasetDistillationComprehensive2023,yangWhatDatasetDistillation2024}.
Early on, it was shown that transforming a teacher (ensemble)'s logits into soft labels can efficiently train a possibly smaller student model~\citep{bucila2006model, bacaruana,hintonDistillingKnowledgeNeural2015}.
This simple mechanism has since been extended to various architectures and modalities in distillation, with numerous methods building on soft label matching as a core ingredient, see e.g.~\citep{Gou_2021,yuDatasetDistillationComprehensive2023,xuSurveyKnowledgeDistillation2024} and references therein. \citet{qinLabelWorthThousand2025a} recently asserted that soft label matching on its own is still competitive for modern vision architectures.
However, despite some theoretical advances \citep{phuongUnderstandingKnowledgeDistillation2019,sagliettiSolvableModelInheriting2020,menonStatisticalPerspectiveDistillation,boix-adseraTheoryModelDistillation2024,dissanayakeQuantifyingKnowledgeDistillation2025}, it still remains unclear exactly what the ``dark knowledge''\citep{hintonDistillingKnowledgeNeural2015} is that soft labels contain, and how to reliably quantify it.\looseness=-1

Among the hypotheses on the regularizing benefits of soft labels~\cite{mullerWhenDoesLabel2020, yuanRevisitingKnowledgeDistillation2021, zhouRethinkingSoftLabels2021a}, one line of reasoning suggests that they are effective because they encode structure reflective of the data distribution~\cite{phuongUnderstandingKnowledgeDistillation2019,menonStatisticalPerspectiveDistillation}.
This implies that soft labels should be especially useful when the data distribution involves low-rank patterns and compresses informative representations to manifolds that allow for generalizing solutions.
These intuitions have been verified in experiments on natural image classification data, where the top-few teacher soft labels play a crucial role in achieving a performance that matches the teacher closely~\cite{qinLabelWorthThousand2025a}.
However, most of current large language and vision models involve not only generalizing skills but also the memorization of facts and associations~\cite{xuSurveyKnowledgeDistillation2024}.
Transferring the full range of a model’s capabilities, including both generalizing patterns and memorized facts, therefore requires an understanding of whether the soft labels convey both types of information.

\begin{figure}
    \centering
    \includegraphics[width=\textwidth]{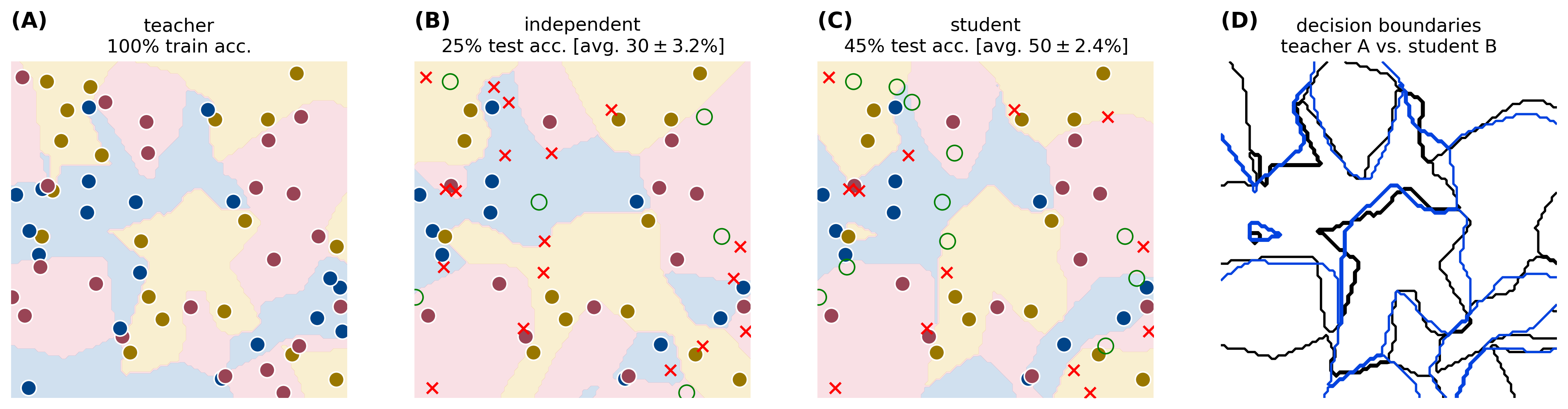}

    \caption{\textbf{Information leakage via soft labels.} We examine fully connected networks with ReLU activations and $p=100$ hidden neurons and biases. A teacher network is trained on 2-dimensional input data $\Dteacher$ with i.i.d. random uniform labels drawn from $\{{\color{RoyalBlue}1},{\color{Maroon}2},{\color{Dandelion}3}\}$. \textbf{(A)} Visualizes the data with the teacher that achieves $100\%$ accuracy. Then, teacher data is partitioned into two disjoint sets $\Dstudenttrain$ and $\Dstudenttest$ at a $(60\%, 40\%)$ ratio. We examine 2 settings: Training student networks via cross-entropy \textbf{(B)} on the class information only, making the student independent from the teacher, and \textbf{(C)} on soft labels obtained from the teacher via softmax on the logits. While the independently trained model only achieves trivial accuracy of $\sim30\%$, students that fit the teacher's soft labels achieve \textit{non-trivial test accuracy} of $\sim50\%$. Markers indicate data from the test set, and whether it was classified {\color{BrickRed} wrongly ($\times$)} or {\color{OliveGreen} correctly ($\circ$)}. We report averages over $5$ initializations with the standard error on the mean. \textbf{(D)} The decision boundaries for teacher \textit{A} (black) and student \textit{B} ({\color{Blue}blue}). In Appendix~\ref{app:visual_example} we show another example for 20 classes.
    \looseness=-1}
    \label{fig:visual-example}
\end{figure}

While there is a long history on understanding the memorization capacity of neural networks theoretically~\cite{hopfield}, practical investigations are more recent~\cite{zhangUnderstandingDeepLearning2017,luScalingLawsFact2024,chenMultiPerspectiveAnalysisMemorization2024}, and there is comparatively little work on how memorized facts can be transferred from white-box teachers with soft labels. Dataset distillation aims to elicit specific knowledge and general skills from the teacher by creating dedicated training data, yet there has not been a clear focus on disentangling how different generalizing and memorizing skills are transferred. Most prior work focuses on generalization and structure, leaving open the question of how memorization behaves during distillation.
To fill this gap, we ask:
\begin{center}
\textit{Do the teacher's soft labels encode memorized knowledge?\\ -- And if yes, can students pick up this non-trivial information?}
\end{center}
To isolate the role of memorization in  distillation with soft labels, we train teacher networks to perfectly fit a finite dataset of input–label pairs. 
We then distill their ``memorized'' knowledge into soft labels, to train students who see only a fraction of those pairs, the distilled dataset, and are evaluated on the held-out remainder. 
We apply this protocol both to (i) small transformers on structured algorithmic tasks and (ii) fully connected networks on random i.i.d. data.
While in (i) we exploit delayed generalization to obtain memorizing teachers, (ii) has no latent structure by design which leads to teacher memorization. Despite its simplicity, the controlled memorization-only setting (ii) has, to our knowledge, not been studied previously in the distillation literature.

For both cases we answer our question \textit{positively}:  From training on the teacher's soft labels a student can indeed learn non-trivial information about held-out memorized data. A simple visual example for knowledge distillation in two dimensions is shown in Fig.~\ref{fig:visual-example}. We summarize our specific contributions below\footnote{Our results and the code to reproduce them are available at {\tiny\url{https://github.com/SPOC-group/dataset-distillation-memorization}}.}:
\begin{itemize} 
    \item We demonstrate for both structured but memorized datasets and purely random i.i.d. data that students trained on teacher's soft labels can consistently recover non-trivial -- in some cases perfect -- accuracy on data the teacher memorized but the student never saw.
    \item  We show that this effect depends strongly on the temperature with which the soft labels are created from the teacher logits and can be interpreted as a regularizer that interpolates between fitting the teacher function and recovering only the ground-truth training labels.
    \item For random i.i.d. data, we show that in logistic regression, simple closed-form capacity and identifiability thresholds separate distinct leakage regimes, and that these thresholds extend to the multi-class case with similar qualitative behavior. For ReLU MLPs, the soft label memorizing and teacher-matching solutions are distinct; the student transitions from the former to the latter only once the teacher is identifiable, with a sudden jump in accuracy.
\end{itemize}

\section{Related Work}
\paragraph{Knowledge distillation.}
Soft labels have been a central component of knowledge distillation since its inception~\cite{bucila2006model,bacaruana,hintonDistillingKnowledgeNeural2015}, and have been applied across a range of domains~\cite{gouKnowledgeDistillationSurvey2021a,xuSurveyKnowledgeDistillation2024}. Prior work has attributed their effectiveness to regularization effects~\cite{mullerWhenDoesLabel2020,yuanRevisitingKnowledgeDistillation2021} or to their ability to encode statistical structure aligned with the data distribution~\cite{menonStatisticalPerspectiveDistillation}. These explanations typically assume that the teacher model reflects meaningful structure in the data. In contrast, we isolate the role of soft labels when the teacher has memorized unstructured data and student and teacher have matched capacity. Under these conditions, the student achieves non-trivial accuracy on held-out memorized examples, suggesting that soft labels can transmit memorized information beyond mere regularization. Moreover, we find that even weak statistical signals in soft labels can suffice to support generalization on memorized data.\looseness=-1\\
Theoretical investigations have considered deep linear networks~\cite{phuongUnderstandingKnowledgeDistillation2019}, learning theoretic analyses based on linear representations~\cite{boix-adseraTheoryModelDistillation2024} and information-theoretic limits on knowledge transfer such as the knowledge retained per neuron~\cite{dissanayakeQuantifyingKnowledgeDistillation2025, zhangQuantifyingKnowledgeDNN2022}.
In a similar setting as ours,  \citet{sagliettiSolvableModelInheriting2020} analyze regularization transfer from teacher to student, but take a teacher as a generating model itself rather than letting it memorize a fixed dataset.

\paragraph{Dataset distillation.} With dataset distillation one aims to construct a small set of synthetic input-output pairs that transfer both generalization ability and task-specific knowledge~\cite{wangDatasetDistillation2020,yuDatasetDistillationComprehensive2023}. These synthetic examples often lie off the natural data manifold and are difficult to interpret, yet they remain effective for training student models~\cite{yangWhatDatasetDistillation2024}. Similar effects have been observed with arbitrary transfer sets~\cite{nayakEffectivenessArbitraryTransfer2020}, where inputs are sampled in a class-balanced manner from the teacher’s domain. These findings suggest that effective distillation may depend less on input realism and more on whether the teacher function can be inferred from the supervision~\cite{cazenavetteGeneralizingDatasetDistillation2023}. While we do not modify the input distribution, our analysis shows that when the data is sufficient to identify the teacher, and softmax temperatures are high, the student can learn the teacher functionally rather than merely class labels.

\paragraph{Memorization.} \citet{zhangUnderstandingDeepLearning2017} famously  showed that deep networks can fit completely random labels, demonstrating their large capacity to memorize arbitrary data. We extend this observation by studying how such memorized information can be transferred via distillation with soft labels.
This is relevant for modern large language models which do not memorize their training corpus, but still require mastering factual recall~\cite{chenMultiPerspectiveAnalysisMemorization2024}.
However, memorizing additional facts incurs a linear cost in model parameters~\cite{luScalingLawsFact2024}.
\citet{bansalMeasuresInformationReflect} distinguish example-level and heuristic memorization, where the latter relies on shortcuts or spurious correlations , which is known to hurt generalization~\cite{bayatPitfallsMemorizationWhen2024}.  In our random data setup, correlations in the dataset arise only from its finiteness, and our analysis in the large data and parameter limit rules out any spurious effects that are not incurred by the teacher.

\section{Notation and Experimental Setting}

\paragraph{Data.} We consider input-output pairs in a classification setting, where input coordinates are $\mathbf x \in \mathbb{R}^d$ and there are $c$ possible labels $y \in \{1,\ldots,c\}$. 
This dataset is available either through the finite set $\mathcal D$ (Section~\ref{sec:mod-addition}) or a generating model from which we can sample i.i.d. (Section~\ref{sec:random-data}).\\
We define a finite dataset of $n$ such samples (elements) from  $\mathcal D$ as $\Dteacher = \{(\mathbf x^\mu, y^\mu)\}_{\mu=1}^n$. To evaluate generalization of the teacher we consider $\Dval$, which is either $\mathcal D \setminus \Dteacher$ or an independent sample.\\
For knowledge distillation the teacher dataset~$\Dteacher$ is randomly partitioned into two disjoint subsets: the student training set~$\Dstudenttrain$ and the student test set~$\Dstudenttest$. We refer to  $\rho = |\Dstudenttrain|/n$ as the student's training data fraction.

\paragraph{Models and Training.}
\newcommand{\temp}{\tau}
\newcommand{\fteacher}{f^\star}
All models we consider are parameterized functions \( f_\theta : \mathbb{R}^d \rightarrow \mathbb{R}^c \) that map inputs $\mathbf x$ to class logits $\mathbf z\in \mathbb R^c$. Predictions are obtained by applying an argmax over the output logits. 
We use the cross entropy loss for supervised classification. For $\mathbf{y} \in \mathbb{R}^c$ being the one-hot encoded label vectors, the cross-entropy loss with temperature $\temp$ is\looseness=-1
\begin{align}
    \mathcal{L}_{\mathrm{CE}}(\{\mathbf{x}^\mu,\mathbf{y}^\mu\}_n) 
    &= -\sum_{i} \sum_{k=1}^c (\mathbf{y}^\mu)_k \log\left[ \sigma_\tau\left(f_\theta\left(\mathbf{x}^\mu\right)\right)_k\right] \,,\\
    \sigma_\tau(\mathbf{z})_k &= \frac{\exp(z_k / \tau)}{\sum_{j=1}^c \exp(z_j / \tau)}\,.\label{eq:softmax}
\end{align}

\begin{figure}
    \centering
     \begin{adjustbox}{valign=c}
        \includegraphics[width=0.35\linewidth]{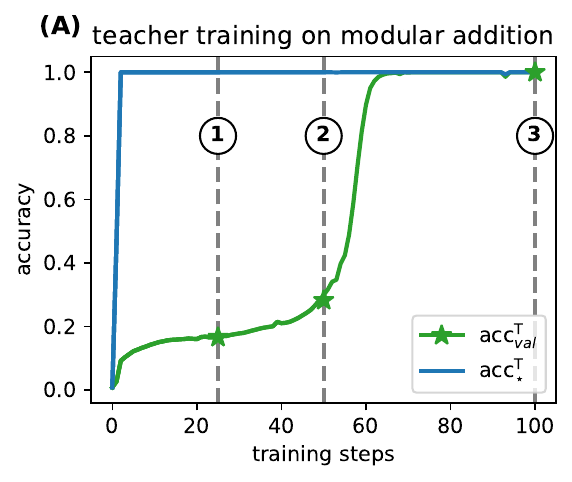}
    \end{adjustbox}
    \hspace{1em}
    \begin{adjustbox}{valign=c}
    \textcolor{gray}{\rule{0.5pt}{0.35\linewidth}}
    \end{adjustbox}
    \hspace{1em}
    \begin{adjustbox}{valign=c}
        \includegraphics[width=0.5\linewidth]{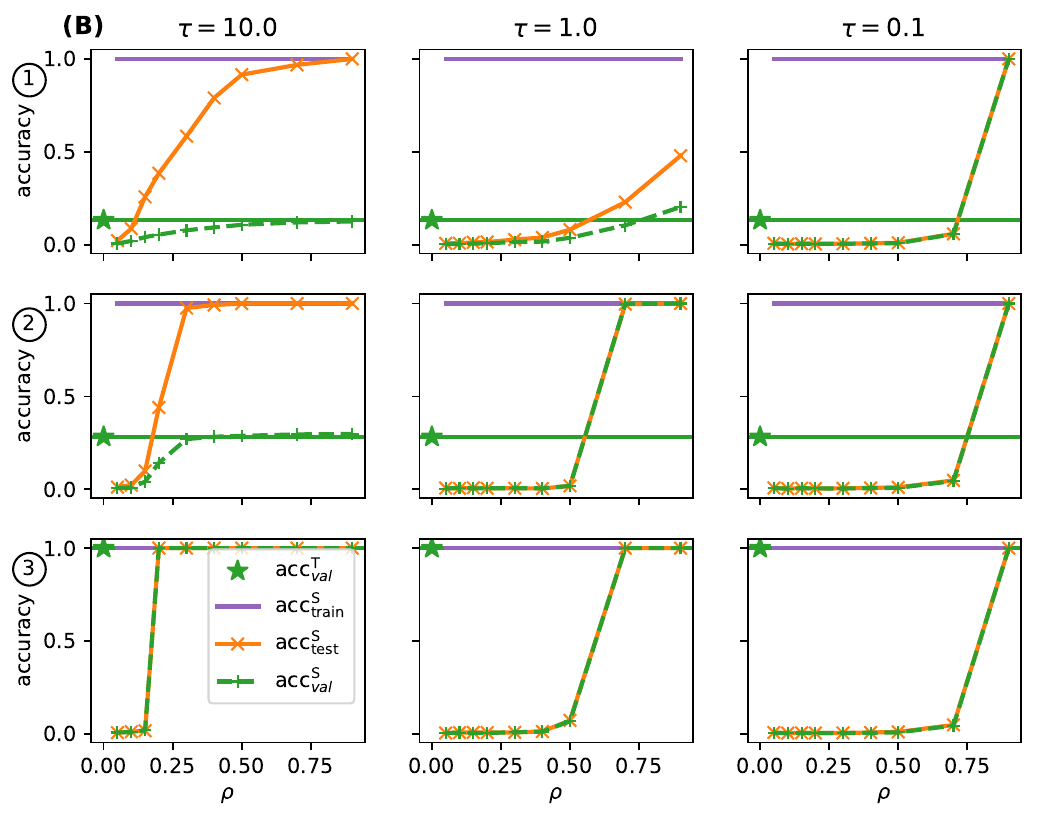}
    \end{adjustbox}
     
    \caption{\textbf{Information leakage via soft labels for structured data in transformers.} \textbf{(A)} Loss curves for small transformers trained on the $30\%$ of the modular addition task with $p=113$. The models \iris{1}, \iris{2} and \iris{3} are stopped after different training times. 
    \textbf{(B)} Students with a matching architecture trained on the respective teachers \textit{(rows)} with different softmax temperatures $\temp$ \textit{(columns)}. We show the students train and test error, and their accuracy on $\Dval$. For comparison, we show the teachers validation accuracy as a horizontal line, marked with a green star. Appendix~\ref{app:transformer_architecture} describes architecture and training details. The same experiment is repeated for ReLU MLPs in Appendix~\ref{app:modular-addition-mlp}.\looseness=-1 }
    \label{fig:modular-addition}
\end{figure}
To transfer the knowledge from a teacher $\fteacher$ to a student $f_\theta$ we train them using the teacher's soft labels. This is achieved using cross-entropy loss, but instead of the ground truth one-hot vector $\mathbf{y}^\mu$ we use a given teacher network's soft labels $\hat{\mathbf{y}}^\mu = \sigma_\temp(\fteacher(\mathbf{x}^\mu))$.
We train using the Adam optimizer~\citep{kingma2014adam} with full batches and default PyTorch settings~\citep{paszke2019pytorchimperativestylehighperformance}. 

\paragraph{Evaluation.} The performance of the teacher and student is measured in terms of accuracy of the argmax over the logit outputs. We distinguish different measurables
\begin{multicols}{2}
\begin{itemize}[left=0pt,labelsep=0.5em,itemsep=0.5em]
  \item $\accTrainTeacher$ -- teacher on $\Dteacher$ (memorization),
  \item $\accTrainStudent$ -- student on $\Dstudenttrain$ (training),
  \item $\accTestStudent$ -- student on $\Dstudenttest$ (test),
  \item $\accValTeacher$ -- teacher on $\Dval$ (T-generalization),
  \item $\accValStudent$ -- student on $\Dval$ (S-generalization).
\end{itemize}
\end{multicols}
When there is no structure in $\mathcal{D}$, the best estimator is random guessing the classes at an accuracy of~$1/c$ for $\accValTeacher$ and $\accValStudent$.

\section{Leaking Held-Out Memories when Data is Structured}\label{sec:mod-addition}

To complement our 2D toy setting from Fig.~\ref{fig:visual-example}, we now study whether the leakage of memorized information through soft labels also occurs in more realistic architectures and structured data.
Specifically, we use the modular addition task and a single layer transformer following the analysis of grokking by~\citet{nanda2023progressmeasuresgrokkingmechanistic}. The authors showed that in this setting training exhibits two phases: Even though the teacher quickly learns to fit the training set, generalization to the task is delayed. This allows us to isolate two different settings: Teachers that memorize their training set without discovering structure, and those that generalize.
From this, we examine how student learning varies based on what the teacher has learned and how this impacts the soft labels leakage of memorized information.

\paragraph{Memorization and generalization in modular addition.} The modular addition task requires adding two integers $a,b \in [0,p]$ modulo $p$. We consider the case where this task is available as a dataset of tuples with one-hot encoded tokens $x=(a,b,p)\in \{0,1\}^{3p}$ with the label $y\in [0,p-1]$. For our experiments we consider only the case where $p=113$, so that the size of the complete data distribution is $|\mathcal D|=113^2=12,769$.
We train the teacher on 30\% of this data, the set $\Dteacher$ ($n=3,830$).
The other part of the original dataset $\mathcal D$ is kept aside for validation in $\Dval$. 
When training a student we split $\Dteacher$ into two disjoint sets $\Dstudenttrain$ and $\Dstudenttest$, where $|\Dstudenttrain| = \rho n$. The set $\Dstudenttest$ tests for the held-out memorized samples.\\
We train transformer architectures with a single layer (see Appendix~\ref{app:transformer_architecture}).
To analyze teachers that have memorized the input data to different degrees, we stop the training at three different points, see Fig.~\ref{fig:modular-addition}(A). Qualitatively we use the teachers generalization on $\Dval$ as a measure of how much structure from the data was already discovered. At \iris{1} the least structure is known, slightly more at \iris{2}, and the teacher completely generalizes at \iris{3}.
In the following, we use these teachers to train students via soft labels. We use different temperatures $\tau$ to generate the soft labels; results as shown in Fig.~\ref{fig:modular-addition}(B).
Importantly, we observe that $\accTrainStudent$ is always 100\%, regardless of $\rho$ and $\temp$.\looseness=-1


\paragraph{Soft labels may leak memorized information in transformers.}
In Fig.~\ref{fig:modular-addition}(B, first row, left), at $\temp=10$ for teacher \iris{1}, we observe that for small $\rho$, i.e. small training sets $\Dstudenttrain$, the student achieves higher $\accTestStudent$ ({\color{Orange}orange}) than $\accValStudent$ ({\color{Green}dashed green}). This means that indeed the soft labels are leaking some information on the training set $\Dteacher$, that accuracy on $\Dstudenttest$ is higher than for $\Dval$.
This indicates that the soft labels leak information specific to the teacher’s training set $\Dteacher$ and allow the student to recover held-out memorized samples, while they do not improve performance on $\Dval$ similarly strongly.
As the fraction of seen teacher data $\rho$ grows, $\accTestStudent$ reaches $1.0$, and $\accValStudent$ approaches $\accValTeacher$.
A similar but more abrupt transition occurs for teacher \iris{2} at the same $\temp=10$ (middle row, left).\\
These results parallel our earlier observations from Fig.~\ref{fig:visual-example}: For some $\Dstudenttrain$ training on the teacher's soft labels leads to non-trivial accuracy on $\Dstudenttest$, which is importantly higher than that on $\Dval$ (analogous to random guessing previously). Unlike the 2D case, however, here the student can \textit{perfectly} generalize to the held-out $\Dstudenttest$.
At the same time, despite $5\times$ longer training than the teacher, at $\tau=10$, these students fail to generalize to $\Dval$ when distilled from the non-generalizing teachers \iris{1} and \iris{2}. Instead they match $\accValTeacher$. This shows that while soft labels can leak memorized inputs, they can also prevent the student from learning latent structure that undertrained memorizing teachers have not discovered.

\paragraph{Higher temperatures are more data efficient for fitting the teacher.} 
At lower temperatures $\tau$, where the soft labels resemble one-hot labels and contain less information about the teacher, the student can outperform the teacher and generalize to $\Dval$. As shown in Fig.\ref{fig:modular-addition}(B, right column), at $\tau=0.1$ student performance even becomes independent of the teacher. The student either fails to generalize due to insufficient data (e.g., at $\rho=0.7$), or exhibits delayed generalization (learning curves Appendix~\ref{app:transformer_speed}).
Only for larger $\temp=10$, learning from the generalizing teacher \iris{3} requires less data with almost immediate generalization on $\Dval$ and for the memorizing teachers \iris{1} \& \iris{2} the students matches their function.
This highlights that higher temperatures both improve data efficiency and convergence speed, and increase the leakage of teacher-specific memorized information.

\section{Distillation for Data without a Latent Structure}\label{sec:random-data}

In the previous section, we used $\accValTeacher$ as a proxy for the amount of teacher memorization. However, a low $\accValTeacher$ does not rule out that the model internally captures some underlying structure, even if it was not predictive.
To isolate memorization in  a controlled setting and to characterize the leakage behavior theoretically, we now consider a data model where there is no structure in the data a priori -- analogous to the introductory example from Fig.~\ref{fig:visual-example}: The entries of the input
$\mathbf x$ are sampled i.i.d. from a Gaussian $x_i \sim \mathcal{N}(0,1)$ and the labels $y \in \{1,\ldots,c\}$ are sampled uniformly and i.i.d. from $c$ classes. The inputs and labels are independent by design, so any teacher needs to memorize the finite dataset $\Dteacher$, failing to generalize to $\Dval$.\\
In the following, we analyze logistic regression, where we can derive closed‑form thresholds for the recovery of $\Dteacher$ in the high-dimensional limit. We consider its multi-class version and show how the same threshold scales in $c$. To estimate the impact of more complex non-linear teachers we analyze leakage in one hidden layer ReLU MLPs.\\
Finally, we show that a teacher GPT-2 model~\cite{Radford2019LanguageMA} fine tuned on a dataset of randomly associated sequences of tokens and classes can also exhibit non-trivial test accuracy $\accTestStudent$ on held-out sequences.

\subsection{Multinomial Logistic Regression}
For multinomial logistic regression we consider linear models $f_{\mathbf W}(\mathbf x) = \mathbf W \cdot \mathbf x$ with $\mathbf W \in \mathbb R^{c\times d}$ that are trained via cross-entropy, known as multinomial logistic regression or softmax regression.
In knowledge distillation this limits us to a setting where teacher and student architecture match.
\begin{figure}
    \centering
    \includegraphics[width=\linewidth]{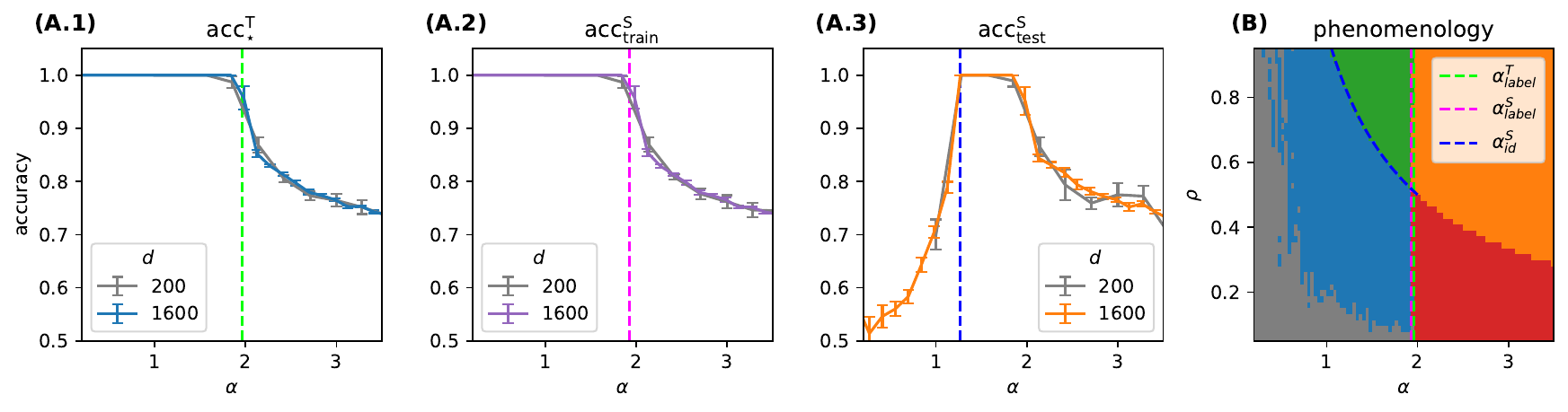}\vspace{-1em}
    \caption{\textbf{Binary logistic regression.} \textbf{(A.1)} We show the training accuracy of the teacher on $\Dteacher$ for $\rho=0.8$ and the students  training \textbf{(A.2)} and  testing accuracies \textbf{(A.3)} of the student on the two partitions of $\Dteacher$. While the teacher is trained via Adam on the logistic loss, the student solutions are obtained from the teacher logits on $\Dstudenttrain$ using the pseudo-inverse. The thresholds $\alphaTlabel$, $\alphaSlabel(\rho)$ and $\alphaSid(\rho)$ are highlighted in {\color{Green} green}, {\color{RubineRed}pink} and {\color{Blue} blue}. \textbf{(B)} depicts the different regimes of teacher/student learning as a function of $\rho$ and the sample complexity $\alpha$. The dimension is fixed at $d=1,600$ and $n$ is varied. We distinguish whether the student fits $\Dstudenttrain$ with $\accTrainStudent\geq0.99$ (gray/blue/green) or not (red/orange). In the regime where it fits the $\Dteacher$, gray implies that the student learns only close to trivial accuracy ($\accTestStudent<0.55$), blue that it is non-trivial ($\accTestStudent<0.99$) and green is perfect ($\accTestStudent\geq0.99$). We measure the MSE loss directly on the teacher logit (see Appendix~\ref{app:mse-logistic}) to evaluate whether the student learned the teacher (orange) or not (red) -- with a threshold set at $0.1$.}
    \label{fig:logictic-c=2}
\end{figure}
\paragraph{Formal analysis: leakage in logistic regression.}  We first consider the case of only two classes, logistic regression\footnote{Here, we exceptionally consider $f:\mathbb R^d \to \mathbb R$ as this is the usual setup of the two class classification problem. The class is then determined by the sign of the single output logit $z$.}. 
When we have direct access to the logit, the problem of recovering the teacher weights $\mathbf W$ under the square loss is equivalent to solving an (over- or under-parameterized) least squares problem, by means of the pseudo-inverse of the input matrix with the logits, i.e., $\widehat{\mathbf{W}} = \mathbf{X}^+ \mathbf{z}$ where $\mathbf{X} \in \mathbb R^{n^s_{\rm train} \times d}$; $\mathbf{z} = f_{\mathbf W}(\mathbf X) \in \mathbb R^{n^s_{\rm train}}$ and $n^s_{\rm train} = |\Dstudenttrain|$.\\
We consider different sample complexities $\alpha = n/d$.
Fig.~\ref{fig:logictic-c=2}(A) shows the accuracy of the teacher on $\Dteacher$, and the train and test accuracies of the student on $\Dstudenttrain$ and $\Dstudenttest$ as a function of $\alpha$ at fixed training set size $\rho = 0.8$. We observe that the $\accTrainTeacher$ and $\accTrainStudent$ start decaying from 1.0 at a given $\alpha$. 
The test accuracy grows monotonically in $\alpha$ from the trivial random guessing accuracy up to perfect accuracy, and at some point it decreases again. 
The general phenomenology concentrates for large $d$ and $n$, as a function of $\rho$, resulting in three thresholds that can be defined in terms of $\alpha=n/d$:
{\setlength{\leftmargini}{6pt}\begin{itemize}
    \item[] $\alpha\leq\alphaTlabel$ --  \textit{teacher memorization capacity}: The teacher can fit all input-class pairs in $\Dteacher$.
    In the proportional limit when $d,n\to\infty$, Cover's Theorem~\cite{coverstheorem} states that $\alphaTlabel \leq 2$ .
    \item[] $a \geq\alphaSid(\rho)$ -- \textit{identifiability threshold}: The student can identify the teacher using the logits, measured through the mean squared error loss on the teacher logits, which occurs at $\alphaSid = 1/\rho$, as the input matrix $\mathbf X$ becomes invertible.
    \item[] $\alpha\leq\alphaSlabel(\rho) $ -- \textit{student memorization capacity}: The student can fit all data from $\Dstudenttrain$ via the input-logit pairs from the teacher.
\end{itemize}}
For finite sizes, we observe that the teacher memorization capacity $\alphaTlabel(d=1600)\simeq1.96$ is already close to the infinite $d$ limit of $\alpha = 2$. 
Beyond this threshold, the student cannot fit $\Dstudenttrain$ perfectly anymore, as it is not memorized by the teacher and information is corrupted.
However, when the teacher does memorize $\Dteacher$ perfectly, the student obtains perfect accuracy on $\Dstudenttrain$ through the logit training set. In this case, we observe that the logits contain a weak signal on the other held-out memorized data and allow the student to obtain $\accTestStudent\geq55\%$ for large enough $\alpha$ and $\rho$, as shown in Fig.~\ref{fig:logictic-c=2}(A.3); some information on the held-out data is leaking.\\
In terms of $\alpha$, $\accTestStudent$ grows monotonically, e.g. for $\alphaSid(\rho=0.8,d=1600)\simeq1.26$, where reaches $\accTestStudent \geq 0.99$ -- even though a fifth of the memorized data that was held-out.
This means that the student can indeed recover the hidden memorized data by recovering the teacher weights $\mathbf W$.

Fig.~\ref{fig:logictic-c=2}(B) shows the different phases can be delineated as a function of $\alpha$ and $\rho$ for a finite fixed $d=1600$: low/no leakage where $\accTestStudent<0.55$, weak leakage of information $\accTestStudent\in(0.55,0.99)$, full recovery of the held-out memorized data $\accTestStudent\geq 0.99$ and failed teacher memorization beyond $\alphaTlabel$. 
We can separate the latter regime into two depending on $\rho$ and $\alpha$, whether the student is able to recover the (non-memorizing) teacher or not, depending on $\mathbf{X}$'s invertibility. 
\begin{figure}
    \centering
    \begin{minipage}{0.24\linewidth}
        \centering
    \includegraphics[width=\linewidth]{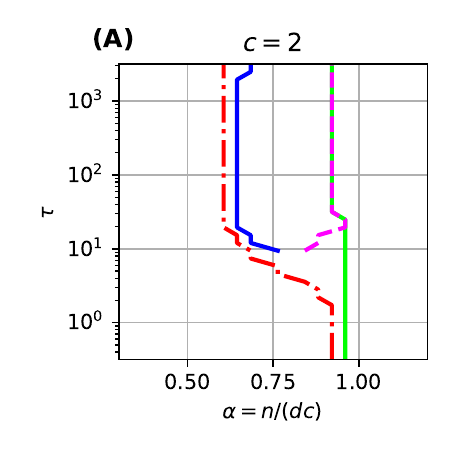} 
    \end{minipage}
    \begin{minipage}{0.73\linewidth}
        \centering
        \includegraphics[width=\linewidth]{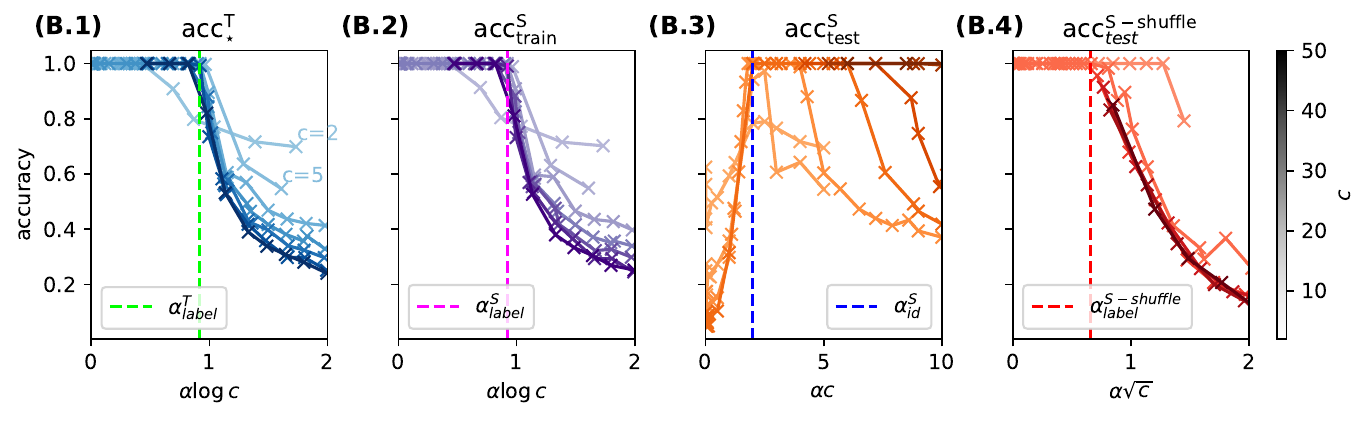}
    \end{minipage}
    \caption{\textbf{Impact of (temperature $\temp$ | number of classes $c$).}
    \textbf{(A)} For the setting with $c=2$ possible classes and $d=1000$, we show the capacity and learning thresholds $\alphaTlabel$ ({\color{Green} green}), $\alphaSlabel$ ({\color{RubineRed}pink}), $\alphaSid$ ({\color{Blue} blue}) and $\alphaSlabelshuffle$ ({\color{Red}red}) as a function of the softmax temperature $\temp$ and the sample complexity $\alpha$; accuracies are reported in Appendix~\ref{app:temperature-visuals}. \textbf{(B)} We take the number of classes as $c=\{2, 5, 10, 20, 30, 40, 50\}$, the larger $c$ the darker the color, and give train and test accuracies in varying scales $\alpha \cdot \{c, \sqrt{c}, \log c\}$. Here $\rho=0.55$ is fixed and $d\in \{100,1000\}$ is varied for computational efficiency depending on $\alpha$ and $\temp=10$.}
    \label{fig:mnr-c-scaling}
\end{figure}

\paragraph{The impact of temperature on memorization.} In practical knowledge distillation with more expressive networks one cannot simply invert but instead one minimizes the cross-entropy loss on soft labels via gradient methods.
Creating soft labels from a students logits requires choosing a temperature $\tau$ in the softmax function~\eqref{eq:softmax}. 
With $\temp \to 0$ one recovers the one hot encodings of the labels and thereby destroys any information that would have been embedded by the teacher. At the other limit, when $\temp \to \infty$, the soft labels become uniform and information about the labels and the teacher is destroyed.\\
For the case of multinomial regression with two classes Fig.~\ref{fig:mnr-c-scaling}(A) shows the relevant thresholds in terms of $\alpha = n/(dc)$ and on the temperature $\temp$ for a fixed $\rho=0.8$ (for accuracies see Appendix~\ref{app:temperature-visuals}).\\
Next to $\alphaTlabel$, $\alphaSid(\rho,\temp)$, and $\alphaSlabel(\rho,\temp)$, we introduce another threshold, $\alphaSlabelshuffle(\rho,\temp)$, derived from a controlled experiment. For each input $\mathbf{x}$ with class $y$ in $\Dstudenttest$, we assign a soft label sampled from a different teacher input $\mathbf{x}'$ within the same class ($y = y'$). This procedure preserves the correct class identity -- the highest soft label entry still corresponds to $y$ -- but removes any teacher-specific information about $\mathbf{x}$. As a result, the student sees noisy supervision: It is class-consistent but the correlation between the rest of the soft label and input is broken. We then define $\alphaSlabelshuffle(\rho,\temp)$ as the point at which this noise prevents the student from learning the class signal.\\
In Fig.~\ref{fig:mnr-c-scaling}(A), we observe that $\alphaSlabelshuffle$ transitions from $\alphaSid$ to $\alphaTlabel$ as $\temp$ increases. This supports interpreting $\temp$ as a hyperparameter that shifts the training objective between fitting soft labels and teacher function (high $\temp$) and recovering class identity (low $\temp$).

\paragraph{Multiple classes $c>2$.}  
As the number of classes increases, the student has a $c$-sized soft label available per training sample, which can contain information about other samples. At the same time, the model size of both teacher and student scales with a factor of $c$.
We observe empirically that the behavior for several classes is consistent with that for two classes: The student can learn non-trivial information about held-out memorized samples and achieve up tp $100\%$ accuracy from the soft labels.\\
In Fig.~\ref{fig:mnr-c-scaling}(B) we observe the scaling behavior of the four relevant thresholds in terms of the number of classes $c$ for a \textit{fixed} $\rho=0.8$ and $\temp=10$, leading to
\begin{align*}
\alphaSid \sim 1 / c\,;\hspace{3em}
\alphaTlabel \sim \alphaSlabel \sim 1 / \log c\,;\hspace{3em}
\alphaSlabelshuffle \sim 1 / \sqrt{c}.
\end{align*}
Naturally, only the scaling of  $\alphaTlabel$ is independent of $\rho$ and $\temp$. Specifically the scaling of $\alphaSlabel$ at $\temp\to0$ should arrive at $\alphaTlabel$. Nonetheless, the order of the thresholds in $\alpha$ remains the same, retaining the original dependence. In Appendix~\ref{app:temperature-visuals} we confirm this for varying temperatures and a fixed $c=10$, where the phenomena are consistent with $c=2$.

\subsection{Two Mechanisms for Leaking Memorized Information in ReLU MLPs}\begin{wrapfigure}{r}{0.6\textwidth}
    \centering
    \vspace{-1em}
    \includegraphics[width=\linewidth]{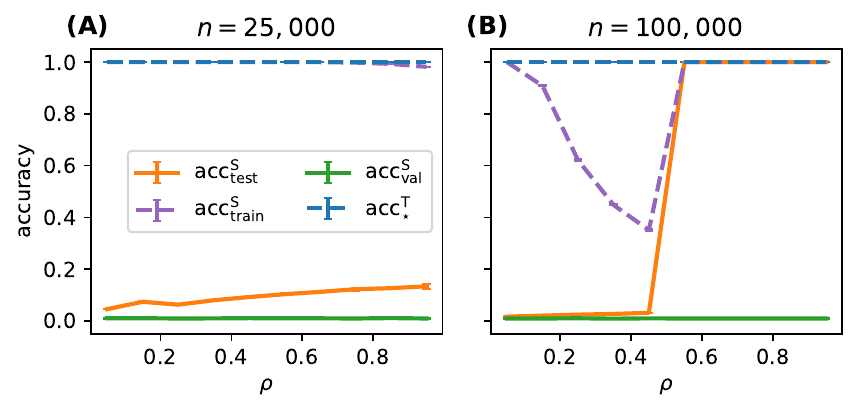}\vspace{-2em}
    \caption{Both teacher and student are MLPs with a single hidden layer of size $p=500$ and ReLU activations.  The inputs are $d=1000$ and $c=100$. The teacher successfully memorizes a training set $\Dteacher$ of size 25,000 \textbf{(A)} and 100,000 \textbf{(B)}. We track the accuracies on both via $\accTrainStudent$ and $\accTestStudent$, with the standard error on the mean reported for 10 runs. In all cases shown here, some information is leaked statistically, allowing the student to surpass trivial performance on data not seen by the teacher ($\accValStudent$), in some cases reaching up to 100\% test accuracy.\looseness=-1}\label{fig:high-dim-example}\vspace{-1em}
\end{wrapfigure}
In this section, we show that ReLU MLPs already exhibit more complex behavior for the same random uncorrelated inputs and labels as before than the multinomial regression case.
In Fig.~\ref{fig:high-dim-example} we consider a matched teacher and student, ReLU MLPs with a single hidden layer, with $c=100$ classes.
On the $x$-axis we vary the fraction $\rho$ of $\Dteacher$ that is observed by the student.
In Fig.~\ref{fig:high-dim-example}(A), the $\accTrainStudent$ and $\accTestStudent$ exhibit a similar phenomenon as before for the logistic regression in Fig.~\ref{fig:logictic-c=2}(D):
While the student memorizes its own training set perfectly, the accuracy  $\accTestStudent$ on the held-out data is non-trivial and increases monotonically as more and more data from $\Dteacher$ is available.
However,  at a higher sample complexity shown in panel (B) of the same figure, we observe two new phenomena:
A first observation is the presence of a phase where $\accTestStudent$ slowly drops while the teacher accuracy $\accTrainTeacher$ remains perfect. Meanwhile, $\accTestStudent$ is \textit{lower} than for the same $\rho$ at lower sample complexity $\alpha$. This is inconsistent with the previous observation, where larger $\Dteacher$ helped identifying the teacher better and therefore led to higher accuracy.
A second observation is the marked jump \textit{after} the drop in $\accTrainStudent$, where both $\accTrainStudent$ and $\accTestStudent$ immediately rise to $100\%$ accuracy.

\begin{figure}[H]
    \centering\vspace{-1em}
    \includegraphics[width=1.0\linewidth]{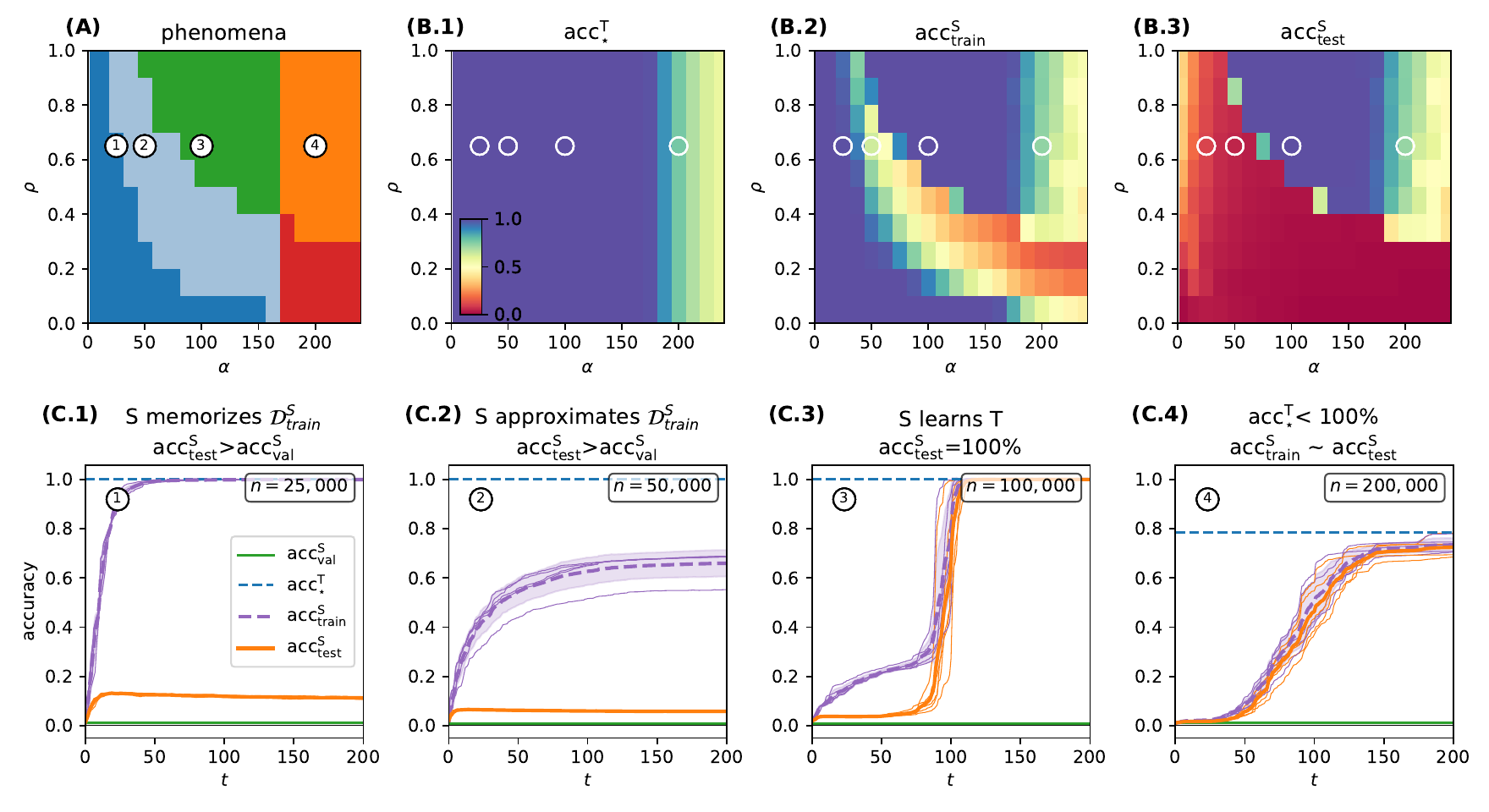}\vspace{-2em}
    \caption{\textbf{Leakage in 1-hidden layer ReLU networks.} The teacher and student architectures match as single hidden layer ReLU networks with $p=500$ for varying settings of sample complexity $\alpha = n/(dc)$ and student training fractions $\rho$. The number of samples $n$ is changed while $c=100$ and $d=1000$ and the temperature $\temp=20$ are fixed. Each experiment is repeated 5 times and average accuracies are reported. \textbf{(A)} Different regimes distinguish the type of generalization the student achieves: ({\color{RoyalBlue}blue}) weakly with memorization of $\Dstudenttrain$; ({\color{Cerulean}light blue}) weakly but without memorization of $\Dstudenttrain$; ({\color{Green}green}) perfectly generalizing to held-out memorized data; ({\color{Orange}orange}) the teacher cannot memorize $\Dteacher$ but the student fits the teacher nonetheless; ({\color{BrickRed}red}) the teacher cannot fit $\Dteacher$ and the student does not discover the teacher either. \textbf{(B)} $\accTrainTeacher, \accTrainStudent$ and $\accTestStudent$. \textbf{(C)} Accuracy as a function of training time $t$ for fixed $\rho = 0.65$ and different sample complexities $\alpha$ as marked with white circles in (A) and (B), varying $n$ and keeping $d=1000$. For comparison, we show averages of $\accTrainTeacher$ and $\accValStudent$ at the end of training as horizontal lines. }
    \label{fig:hidden-overview}
\end{figure}

\paragraph{Memorization fails before teacher identification succeeds: $\alphaSlabel < \alphaSid$.} To understand these phenomena better, we turn to a more complete picture of the phase space in Fig.~\ref{fig:hidden-overview}(A).
Next to the regions already identified for the logistic regression in Fig.~\ref{fig:logictic-c=2}(B), we split the regions where a weak leakage is detected into two parts: The one where the student perfectly learns $\Dstudenttrain$ and the one where the student does not memorize the training data.
In Fig.~\ref{fig:hidden-overview}(B.2) it is further visible that $\accTestStudent$ decreases before it increases as a function of the sample complexity $\alpha$. 
To understand this behavior we observe $\accTrainStudent$ and $\accTestStudent$ as functions of training time in Fig.~\ref{fig:hidden-overview}(C.3). 
There, a sudden jump in train and test accuracy occurs as a function of student training time at around $t\sim100$. While before the jump, the training and testing accuracy are at different levels (and already non-trivial for the student), they jointly jump to $100\%$ accuracy.
In Appendix~\ref{app:loss-curve-comparison} it is shown that this jump coincides with a drop in the CE loss on the teacher distribution and that just before the transition, $\accTrainStudent$ approaches that of a student trained on intra-class sampled soft labels. This suggests that for ReLU MLPs, there may be two distinct weight configurations: One where the student memorizes the soft labels, and another where it functionally matches the teacher. This distinction was not present for multinomial logistic regression.

\newpage

\vspace{-0.5em}
\paragraph{Memorizing the soft labels vs. generalizing on the teacher function.}
These observations suggest that the student can learn two functionally different solutions that both leak information about held-out memorized data, but differently: One solution memorizes the teacher's soft labels representing $\Dstudenttrain$, and another generalizing solution matches the teacher functionally.
This extends the picture from the multinomial regression, in that not only weakly (and fully) learning the teacher function leads to non-trivial leakage on the held-out set, but also a solution that truly memorizes the soft labels can capture some additional structure on held-out data.
Whether one or the other solution is learned depends non-trivially on the respective capacity thresholds, the algorithm, and the ratio between the teacher and student capacity. In Appendix~\ref{app:ablations} we provide some additional ablations that explore how increasing the parameters via the hidden layer size $p$ and the relative capacity of teacher and student in an unmatched setting impact $\accTestStudent$.
\vspace{-0.5em}
\paragraph{Localizing the information in the soft labels.} In Appendix~\ref{app:ablations} we test the effect of removing an input class $c_i$ from $\Dstudenttrain$ and removing it from the soft labels by zeroing it out for all other classes $c_j \neq c_i$.
We find that while removing the inputs can still lead to a non-trivial accuracy on $c_i$ in $\Dstudenttest$, removing the corresponding soft label entries is detrimental for test performance.
Likewise, zeroing out the smallest $k$ values in every soft label negatively affects $\accTestStudent$.
This leads us to hypothesize that the common practice of using only the top-$k$ largest values may not allow for generalizing on the memorized information.

\subsection{Dataset Distillation for Finetuned GPT-2 Classifiers on Random Sequences}

In order to test whether these phenomena extend to random sequence data, we examine a similar setting with a GPT-2 architecture~\cite{Radford2019LanguageMA}. We consider sequences $x = $\texttt{`429\_3507\_345'}, where each sequence concatenates three random numbers sampled uniformly and i.i.d. between $1$ and $1000$, with a random class $y$ out of $1000$ possible classes.
In our setting $\Dteacher$ contains $6000$ samples of such sequences and their classes.
We equip the next-token prediction backbone GPT-2 with a linear classifier head. We use the standard tokenizer and train the teacher on $\Dteacher$ for $100$ epochs using AdamW with a learning rate of $5 \times 10^{-4}$. \\\begin{wrapfigure}{r}{0.49\textwidth}
    \centering\vspace{-1em}
    \includegraphics[width=0.48\textwidth]{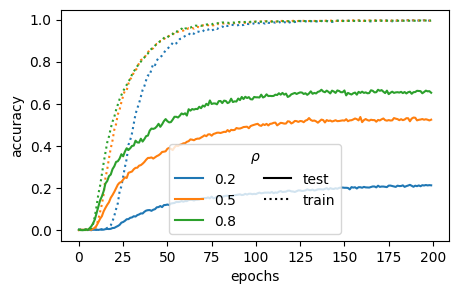}\vspace{-1em}
    \caption{\textbf{Leakage in a large language model for synthetic data.} 
    $\accTrainStudent$ and $\accTestStudent$ for GPT-2 classifier students trained using different $\rho$ fractions of random sentences memorized by a teacher with the same architecture (both pre-trained).
    The size of the memorized training set is $n=6000$ sentences made of three random numbers up to $1000$, each with one of $1000$ classes assigned randomly.
    }
    \label{fig:llmrandom}
\end{wrapfigure}
After successful training, when the teacher memorizes the sentences with $XX\%$ accuracy, we extract the teacher's logits for its training data and create soft labels, with temperature $\tau =20$. 
 We train different the students for different fractions $\rho = \{0.2,0.5,0.8\}$ for $200$ epochs, but otherwise use the same settings as for the teacher. After convergence all three students reach $\accTrainStudent\simeq99.5\%$ and $\accTestStudent=\{0.213,0.524,0.652\}$ (Fig.~\ref{fig:llmrandom}) -- while the test accuracy of random guessing is approximately $\accValStudent=0.1\%$.
For seeing only $80\%$ of the teachers data, the student achieves $>60\%$ accuracy on the held-out data.
This suggests that, similar to single-layer models, an over-parameterized language model may recover a non-trivial fraction of the teacher's held-out memorized data. 
However, despite some exploration of different parameters, we did not yet observe a setting where the teacher function is exactly recovered as for the MLPs, i.e. where the student reaches $\accTestStudent=100\%$.

\section{Conclusion and Discussion}
In this work, we study how memorized data influences distillation. We analyze both  structured data with teachers at varying levels of memorization or generalization as well as teachers that memorized data without a latent structure. We show that students can acquire information about memorized teacher data that was held out during their training from the teachers soft labels. By evaluating performances across teacher and student, we identify distinct regimes: In some, the student memorizes soft labels via statistical leakage; in others, it generalizes the teachers function. Our findings serve as a proof of concept to understand that memorized information can be transferred between models.

\paragraph{Limitations.} Our analysis is restricted to synthetic datasets that are either structured or explicitly memorized and do not capture all aspects of natural data distributions. While this setup allows for precise control and analysis, it limits the immediate application of our findings to real-world tasks. Additionally, we focus on simple models such as logistic regression, single-layer ReLU MLPs and small transformers to enable theoretical insight and controlled empirical study. Even though we show that leakage also occurs for one instance of a large language model with synthetic data, it remains unclear to what extent the identified leakage regimes and thresholds translate to deeper architectures. In particular, we did not explore the role of regularization and optimization that may influence the models capacity. Finally, we only consider memorization and soft labels, which excludes broader knowledge and dataset distillation settings where the teacher jointly learns generalizing structure and memorized data.

\paragraph{Future work.}
On the theoretical side, our framework motivates a more fine-grained analysis of multinomial logistic regression and single layer networks under knowledge distillation, to identify information-theoretic and algorithmic thresholds of fitting structure and memorized data jointly. On the practical side, it is important to understand how models represent both memorized and generalized content when trained on data distributions that mix random and structured information, and in particular how this knowledge can be transferred for efficient dataset distillation, or whether it can be hidden for privacy reasons.

\section*{Acknowledgments}
We thank Luca Biggio for insightful discussions.
This work was supported by the Swiss National Science Foundation under grants SNSF SMArtNet (grant number 212049).


\newpage
\appendix

\section{Visual Example for \texorpdfstring{$d=2$}{d=2} and \texorpdfstring{$c=20$}{c=20}}\label{app:visual_example}

\begin{figure}[H]
    \centering
    \includegraphics[width=\textwidth]{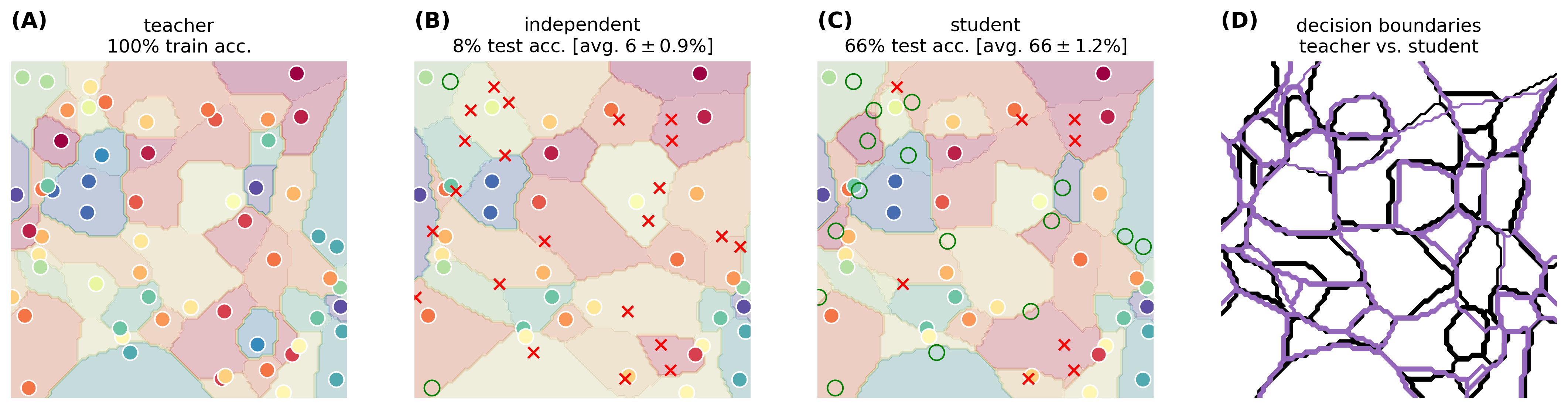}

    \caption{\textbf{Information leakage via soft labels for $c=20$.} We examine fully connected networks with ReLU activations and $p=300$ hidden neurons. A teacher network is trained on 2-dimensional input data $\Dteacher$ with i.i.d. random uniform labels drawn from $20$ classes visualized in a spectrum of colors. \textbf{(A)} Visualizes the random data and teacher decision boundaries, on which the teacher achieves $100\%$ accuracy. Then, teacher data is partitioned into two disjoint sets $\Dstudenttrain$ and $\Dstudenttest$ at $(60\%, 40\%)$ ratios. We examine 2 settings: Training student networks via cross-entropy \textbf{(B)} on the class information only, making the student independent from the teacher,  and \textbf{(C)} on soft labels obtained from the teacher via softmax on the logits with temperature $\temp=20$. While the independently trained model only achieves close to trivial accuracy of $\sim 6\%$, students that fit the teacher's soft labels achieve \textit{non-trivial test accuracy} of $\sim 66\%$. {\color{BrickRed} Red} and {\color{OliveGreen} green} indicate data from the test set, and whether it was classified {\color{BrickRed} wrongly} or {\color{OliveGreen} correctly}. The average test accuracy over 5 initializations is given along with the standard error on the mean. \textbf{(D)} The decision boundaries between teacher (black) and student (purple) correspond very well. }
    \label{app::fig:visual-example}
\end{figure}

As another example that can be visualized analogously to Fig.~\ref{fig:visual-example}, we show a random dataset again in two dimensions but with 20 classes in Fig.~\ref{app::fig:visual-example}.

\section{Supplementary Material for Modular Addition}
\subsection{Implementation details for the transformer}\label{app:transformer_architecture}

As an architecture we consider the single layer transformer from \citet{nanda2023progressmeasuresgrokkingmechanistic}. 
It embeds the $p+1$ tokens into $128$ dimensions. There is a dot-product attention layer with $4$ heads followed by a ReLU MLP with a single hidden layer of $4\cdot 128$  dimensions. A readout layer maps its outputs to the $p$ classes.\\
By design the prediction is autoregressive, but in this case only the last predicted token is relevant and included in the training loss. It becomes a parameterized function $f: \mathbb R^{3(p+1)} \to \mathbb R^{p}$ where $c=p$ and $d=3p$.\\
During training we use weight decay set to $1.0$ and use full batches, which recovers the setting in which grokking was observed~\cite{nanda2023progressmeasuresgrokkingmechanistic}. We train using the Adam optimizer and a learning rate of $0.001$.

\subsection{Generalization speed}\label{app:transformer_speed}

\begin{figure}[H]
    \centering
    \includegraphics[width=0.7\linewidth]{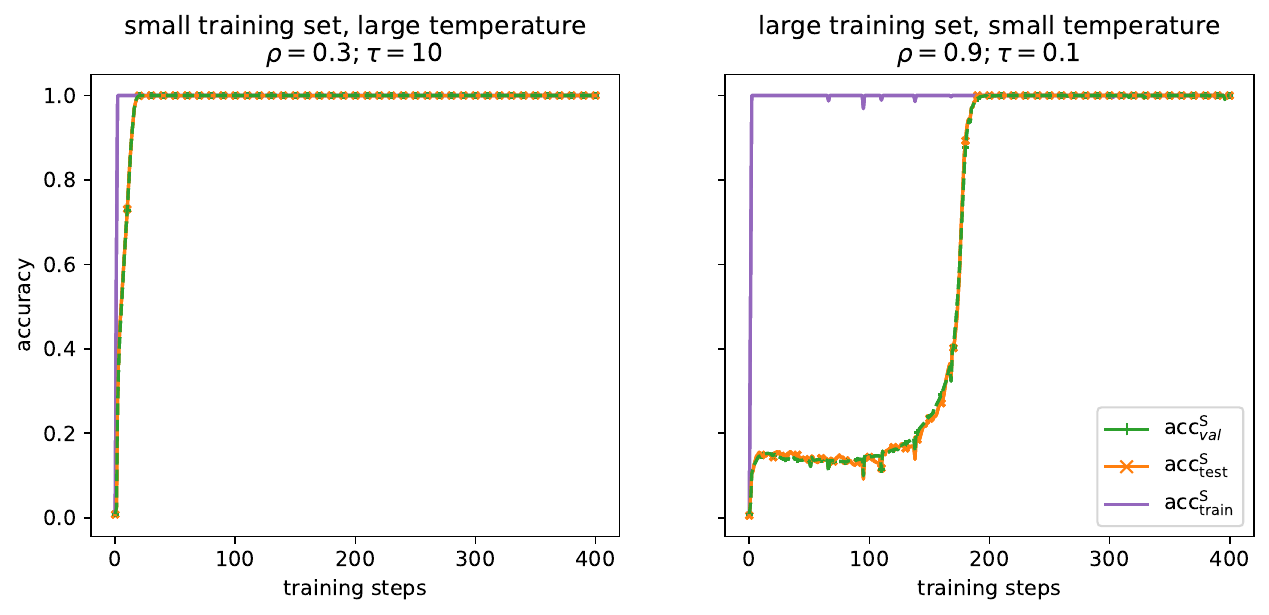}
    \caption{Comparison of the training, test and validation accuracy for transformer students (training and test data were seen by the teacher), for a single run from data used in Fig.~\ref{fig:modular-addition}(A). \textit{(Left)} The student learning from a small training set ($\rho=0.3$) with a high temperature ($\temp=10$) not only memorizes its training data fast but generalizes on the held-out teacher train set and validation set after only few epochs. \textit{(Right)} A student that sees a large training set at a smaller temperature however, exhibits grokking in a similar fashion as the original teacher (see Fig.~\ref{fig:modular-addition}(A)~).}
    \label{fig:gen-speed}
\end{figure}

We examine how fast (in terms of training time) a student reaches perfect accuracy when trained on the soft labels from a perfectly generalizing teacher. In Fig.~\ref{fig:gen-speed} we show exemplary learning curves for the results from Fig.~\ref{fig:modular-addition} with teacher \iris{3} in the main Section~\ref{sec:mod-addition}.

\subsection{Modular addition with MLPs}\label{app:modular-addition-mlp}

\begin{figure}[H]
    \centering
     \begin{adjustbox}{valign=c}
        \includegraphics[width=0.35\linewidth]{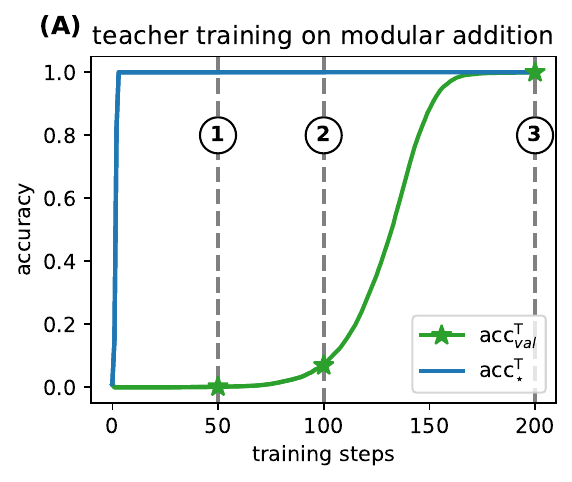}
    \end{adjustbox}
    \hspace{1em}
    \begin{adjustbox}{valign=c}
    \textcolor{gray}{\rule{0.5pt}{0.4\linewidth}}
    \end{adjustbox}
    \hspace{1em}
    \begin{adjustbox}{valign=c}
        \includegraphics[width=0.55\linewidth]{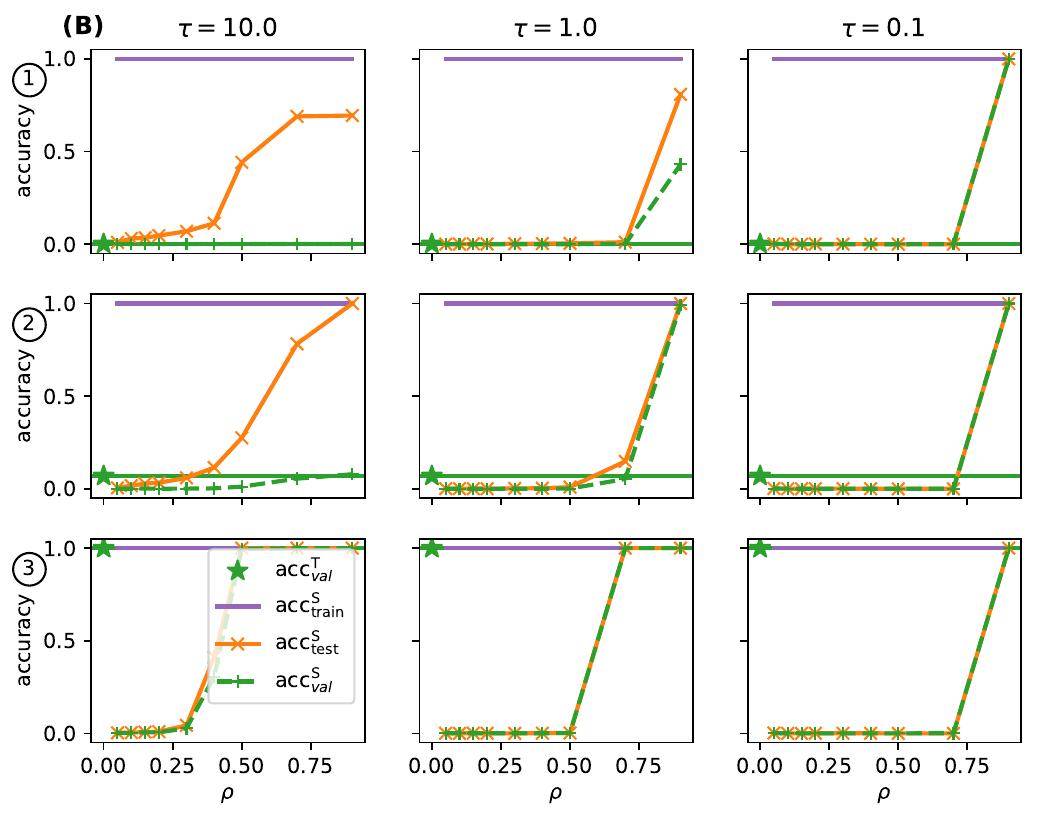}
    \end{adjustbox}
    \caption{Same experiments as in Fig.~\ref{fig:modular-addition}, but with teacher and student architectures that are 2-hidden layer ReLU MLPs. \textbf{(A)} Teacher training that was stopped at different points in training, leading to the teachers \iris{1}, \iris{2} and \iris{3}. \textbf{(B)} Accuracies after training from the student for learning from the different teachers, at different $\rho$ and $\temp$.}
    \label{fig:modular-addition-mlp}
\end{figure}

In addition to small transformers, we repeat the experiment from the main text Fig.~\ref{fig:modular-addition} with 2-hidden layer ReLU MLPs with $200$ hidden neurons each in Fig.~\ref{fig:modular-addition-mlp}.

We also stop the models training at three different points \iris{1}, \iris{2} and \iris{3}, which exhibit almost none, very weak and good generalization.
In analogy to the transformer, at low temperatures the student's behavior becomes independent of the time at which the teacher training was stopped.
Also, for higher temperatures and memorizing students, the $\accTestStudent$ reaches higher values than $\accValStudent$, indicating that the soft labels only transfer information about the memorized labels but not the tasks structure and even prevent the students from generalizing.

\section{Supplementary Material for Random Data}

\subsection{Logistic Regression: Training Students from the Logit}\label{app:mse-logistic}

We define the mean squared error between the teacher \( f^T \) and the student \( f^S \) on the dataset \( \mathcal{D} = \{ \mathbf{x}^\mu, y^\mu \}_{\mu=1}^n \) as:
\begin{align}
\mathrm{mse}(f^T, f^S, \mathcal{D}) = \frac{1}{n} \sum_{\mu=1}^n \left( f^S(\mathbf{x}^\mu) - f^T(\mathbf{x}^\mu) \right)^2.
\end{align}

Let the matching accuracy of the student with respect to the teacher on the student test set be
\begin{align}
\mathrm{acc}^{\mathrm{S}}_{\mathrm{match\text{-}T}} = \frac{1}{n} \sum_{\mu=1}^n \mathbf{1}\left[ \arg\max_j f^S_j(\mathbf{x}^\mu) = \arg\max_j f^T_j(\mathbf{x}^\mu) \right],
\end{align}
where \( f^T_j(\mathbf{x}) \) and \( f^S_j(\mathbf{x}) \) denote the logits assigned to class \( j \) by the teacher and student, respectively.

\begin{figure}[h!]
    \centering
    \includegraphics[width=0.9\linewidth]{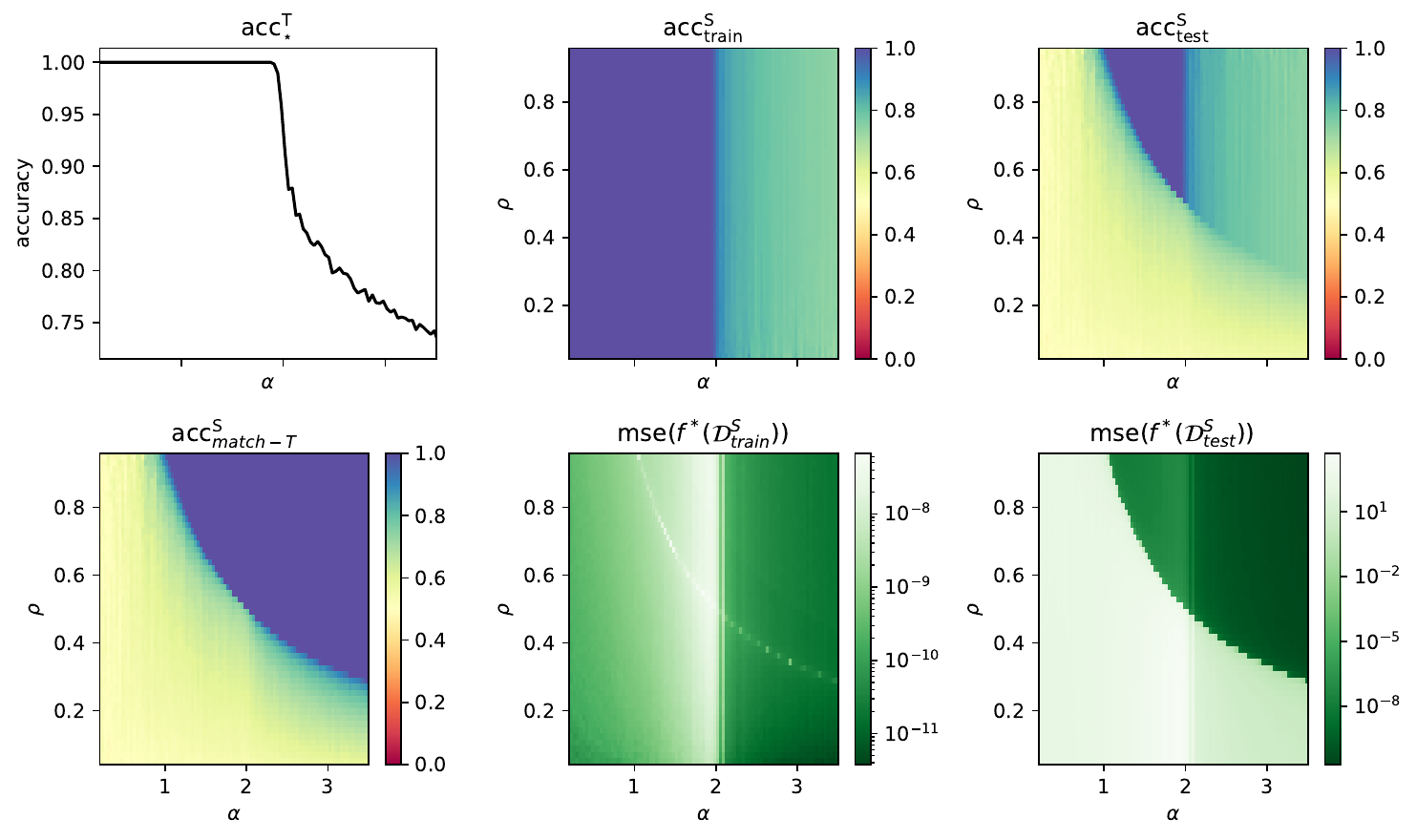}
    \caption{We train logistic regression teacher to fit $\Dteacher$, and recover a student via the teacher logit via a pseudo-inverse. We change the number of samples and keep $d=1600$ while we change $\alpha = n/d$. Every point reports the mean of 5 experiments with different teacher, student and model initializations. For training the teacher we used Adam, learning rate $0.001$ and for $10,000$ steps.}
    \label{app:fig:logistic-regression-phase-diagram-measurement}
\end{figure}

Fig.~\ref{app:fig:logistic-regression-phase-diagram-measurement} shows the different values of $\accTrainTeacher, \accTrainStudent, \accTestStudent,\mathrm{acc}^{\mathrm{S}}_{\mathrm{match\text{-}T}},\mathrm{mse}(f^T, f^S, \mathcal{D}^S_{\text{train}})$ and $\mathrm{mse}(f^T, f^S, \mathcal{D}^S_{\text{test}})$
for $\rho$ and $\alpha=n/d$.

\newpage
\subsection{Varying the Temperature for Multinomial Logistic Regression}\label{app:temperature-visuals}

\begin{figure}[h!]
    \centering
    \includegraphics[width=0.9\linewidth]{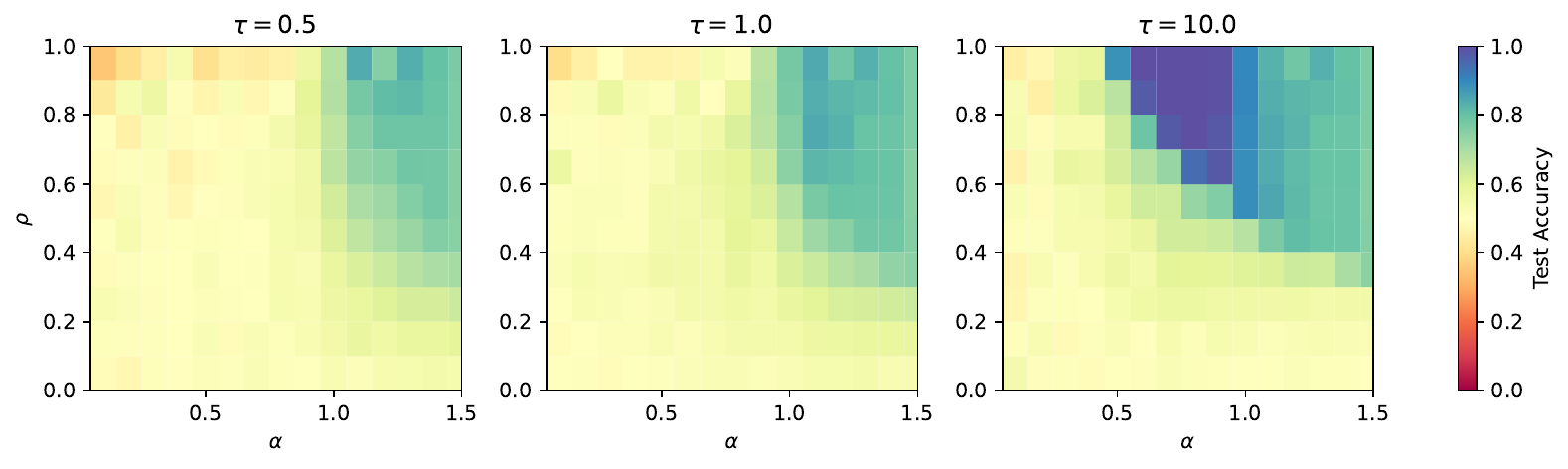}
    \caption{For multinomial regression with two classes and $\alpha=n/(dc)$, and $d=1000$ we report the impact of different temperature in the softmax on $\accTestStudent$. Experiments are repeated 5 times. We train the teacher with Adam and learning rate $0.0001$ for $1000$ epochs and the student with learning rate $0.001$ for $5000$ steps. }
    \label{app:fig:c=2-vary-temp}
\end{figure}

\begin{figure}[h!]
\centering\includegraphics[width=0.85\linewidth]{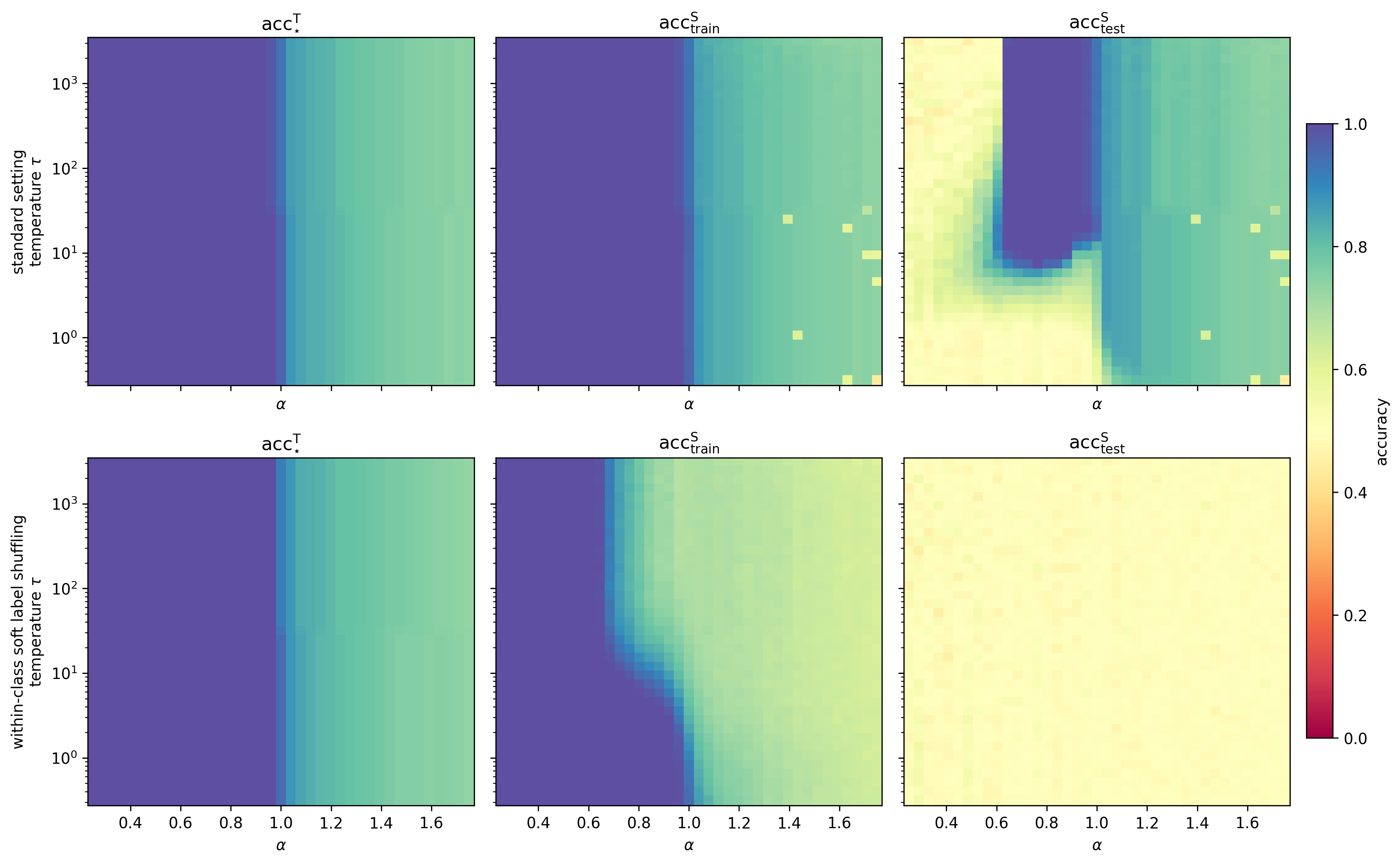}
    \caption{For multinomial regression with two classes $\rho=0.8$ and $d=1000$ we report accuracies for teacher and training data for two experimental settings. The top row is our standard setting and the bottom row re-shuffles the input-soft label assignment within the classes in $\Dstudenttrain$.}
    \label{app:fig:phase-diagram-c=2-measurements}
\end{figure}

\begin{figure}[h!]
    \centering
    \includegraphics[width=0.72\linewidth]{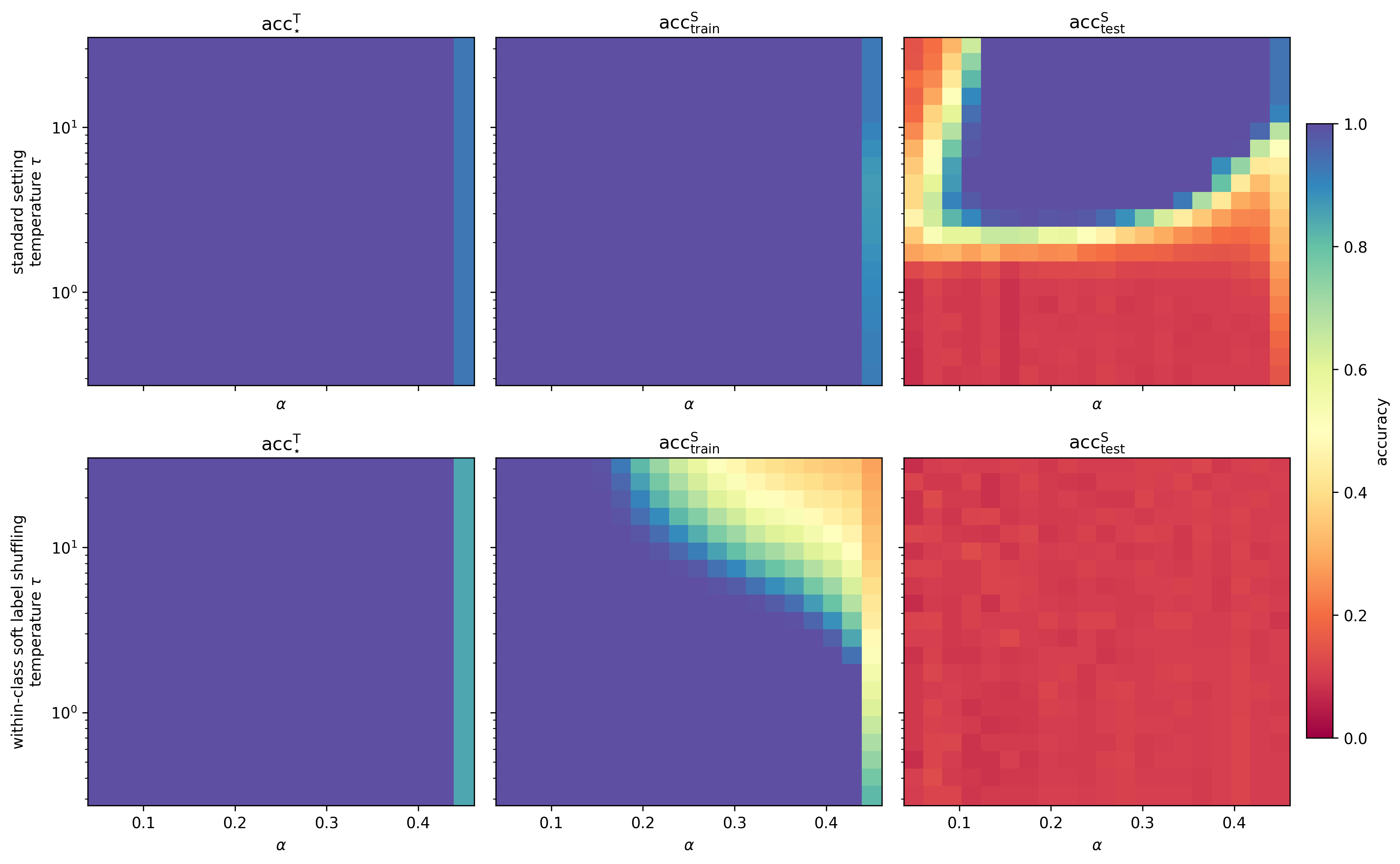}\includegraphics[width=0.25\linewidth]{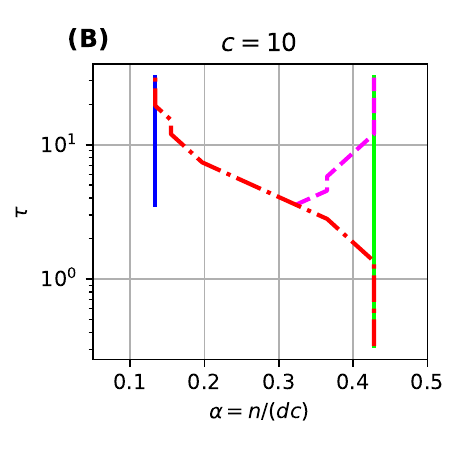}
    
    \caption{Same setting as Fig.~\ref{app:fig:phase-diagram-c=2-measurements}, but with $c=10$ classes instead of $2$. The lines in (B) are the thresholds $\alphaTlabel$ ({\color{Green} green}), , $\alphaSlabel$ ({\color{RubineRed}pink}), $\alphaSid$ ({\color{Blue} blue}) and $\alphaSlabelshuffle$ ({\color{Red}red}) as a function of the softmax temperature $\temp$ and the sample complexity $\alpha$.}
    \label{app:fig:c=10-temperature}
\end{figure}

In Fig.~\ref{app:fig:c=2-vary-temp} we show that the temperature influences whether or not held-out teacher data is leaked to the student for different $\rho$ and $\alpha$ in multinomial logistic regression with $c=2$.\\
In addition, the original measurements that lead to the phase diagram for $c=2$ in Fig.~\ref{fig:mnr-c-scaling}(A) are shown in Fig.~\ref{app:fig:phase-diagram-c=2-measurements}.
We show the same experiment for $c=10$ in Fig.~\ref{app:fig:c=10-temperature}.

\newpage
\subsection{Supplementary training information}\label{app:loss-curve-comparison}

\begin{figure}[h]
    \centering
    \includegraphics[width=0.9\linewidth]{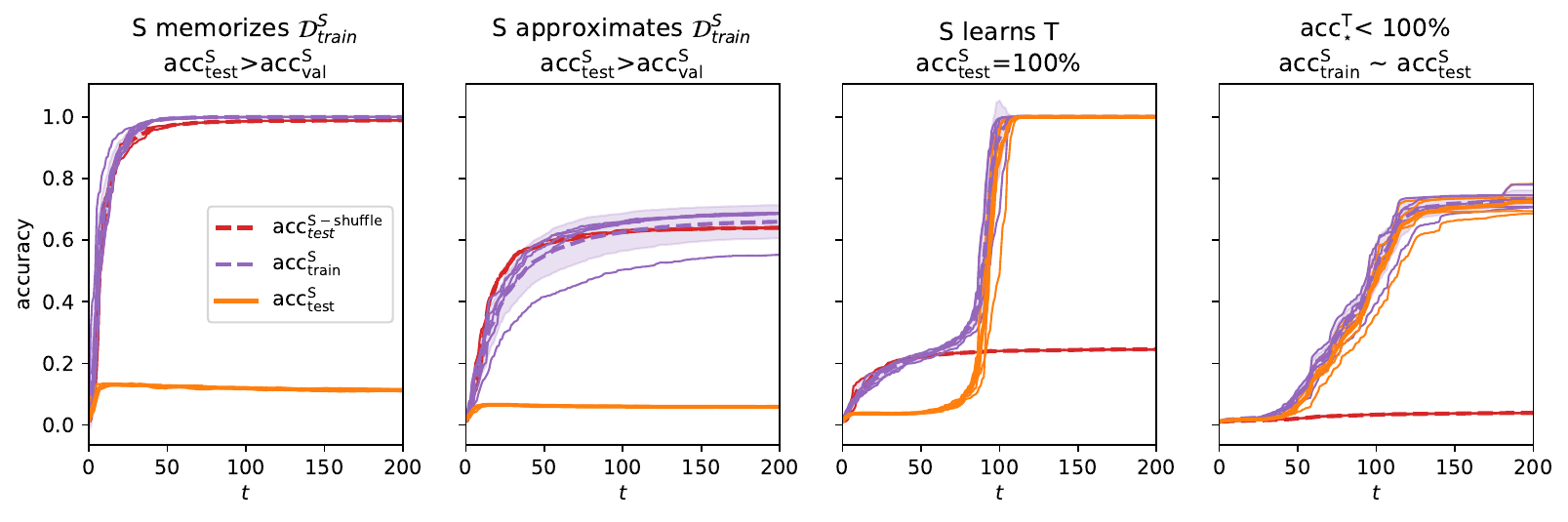}
    \caption{Accuracies for ReLU MLPs from Fig.~\ref{fig:hidden-overview}(C), but this time compared with runs where the input data was shuffled within classes, in red.  Again, the input dimension is $d=1000$ with $c=100$ classes and from left to right the teacher saw $n = 10^3 \times \{15,50,100,200\}$ samples, of which the student was trained with a $\rho=0.65$ fraction. There are 5 runs for the normal student, and 2 for the student that receives the altered teacher data for training.}
    \label{fig:compare-shuffle-mlp}
\end{figure}

\begin{figure}[h]
    \centering
    \includegraphics[width=0.9\linewidth]{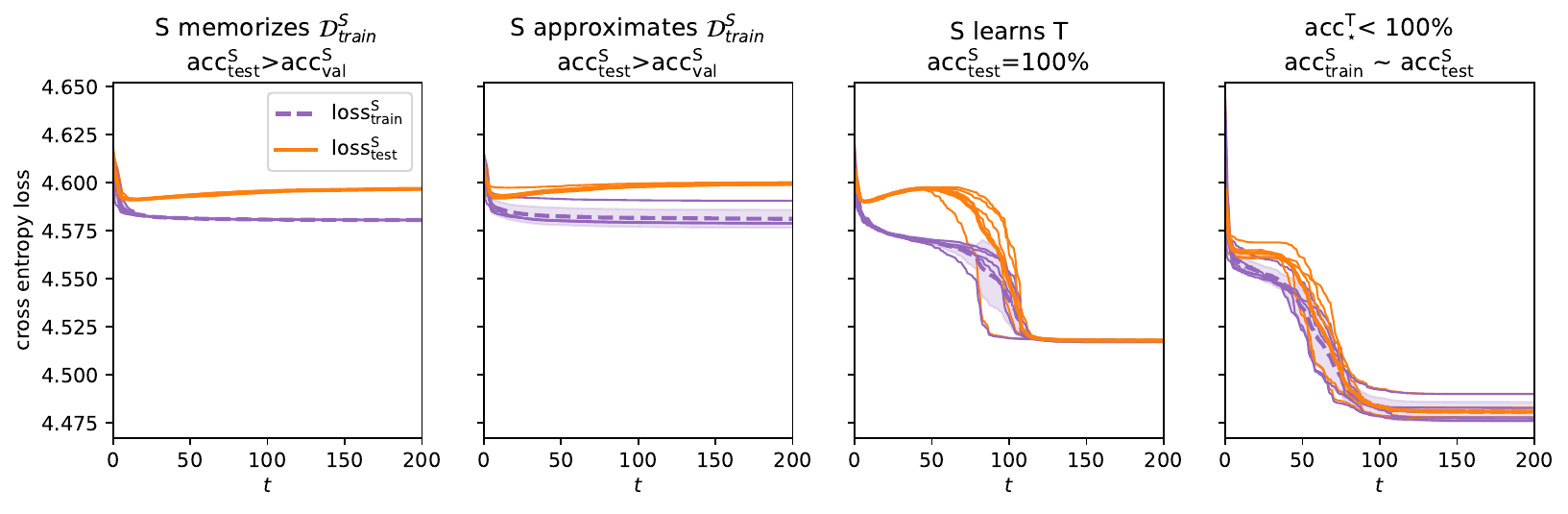}
    \caption{Cross entropy loss on $\Dstudenttrain$ and $\Dstudenttest$ for the results from Fig.~\ref{fig:compare-shuffle-mlp}(C). The input dimension is $d=1000$ with $c=100$ classes and from left to right the teacher saw $n = 10^3 \times \{15,50,100,200\}$ samples, of which the student was trained with a $\rho=0.65$ fraction. We show the average (wide line) and single runs (thin lines) for 5 runs each.}
    \label{fig:losses-mlp}
\end{figure}

We supplement the accuracy curves from Fig.~\ref{fig:compare-shuffle-mlp} for the single hidden layer ReLU MLP with the corresponding losses in Fig.~\ref{fig:losses-mlp}, for different sample complexities $\alpha=n/(dc)$. It is visible, that around the same moment where the jump in accuracy occurs for  Fig.~\ref{fig:compare-shuffle-mlp}(C.3), the loss also drops significantly.

 In Fig.~\ref{fig:compare-shuffle-mlp} we compare the learning curves from the students in Fig.~\ref{fig:hidden-overview}(C) with those from the students which trains on the within-class shuffled soft labels. In the third panel it is visible, that before the student generalizes to the teacher function, it obtains the same accuracy as the shuffled student. Before this third panel, the accuracy is matching well, whereas for the forth panel the difference comes about quickly. This supports the intuition that the student first attempts to truly memorize, and then generalizes the teacher structure in the third panel.

\newpage
\subsection{Ablations  for ReLU MLPs}\label{app:ablations}

\subsubsection{Varying the soft label content}
In this section we conduct several ablations, to understand which information in the soft labels is crucial for obtaining a good $\accTestStudent$. We conduct three experiments, where we:
\begin{itemize}
    \item Remove small soft label entries: We zero out the entries of the smallest $k$ values of the $c$ soft labels for a given input. This leads to the rest of the vector not summing to one anymore, but the cross-entropy loss can still be computed.
    \item Remove a single class from the training data: We remove the class $c$ from the training data, and evaluate on the test data.
    \item Remove a single class from other classes soft label vectors: We specifically zero out the class $c$ value in the teacher soft labels for all classes $c' \neq c$.
\end{itemize}

\begin{figure}[h]
    \centering
    \includegraphics[width=0.8\linewidth]{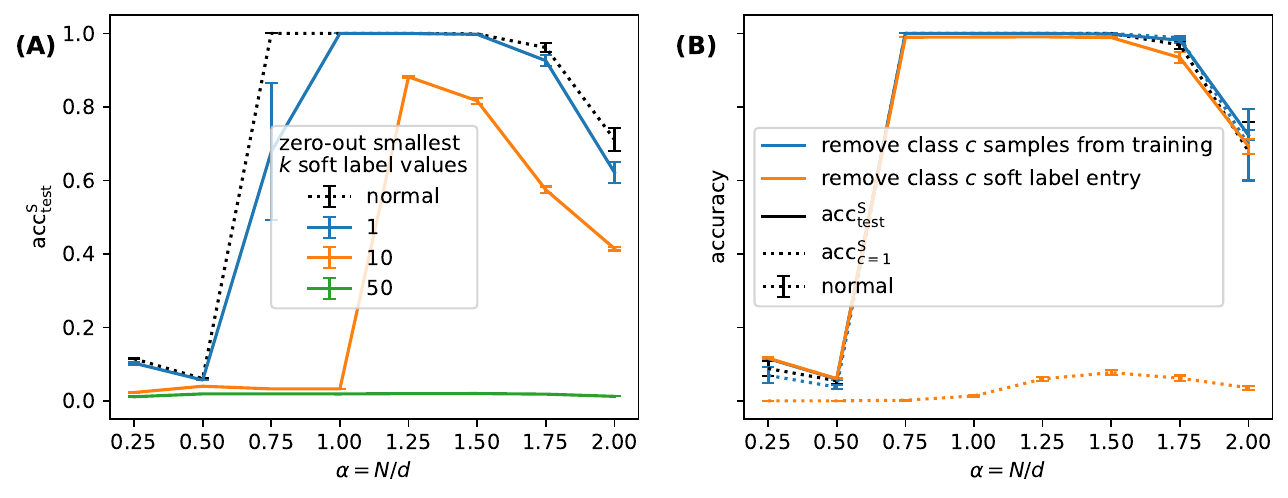}
    \caption{We consider ReLU MLPs with a single hidden layer where student and teacher have $p=500$. We keep $d=1000$, $c=100$ and the fraction on which the student is trained is $\rho=0.7$, and we use a temperature of $\tau=20$.  \textbf{(A)} We compare  zeroing out the smallest $k \in \{1,10,50\}$ out of the 100 values from the soft label vector in the training data with a student trained on the unaltered data. \textbf{(B)} We also compare removing samples with class $c$ (here $c=1$ w.l.o.g.) from the training data completely, removing the soft labels of all training data $\Dstudenttrain$ whenever the true labels is not $c$. Every point is the average of 5 random datasets and initializations, with the standard error on the mean shown as bars.}
    \label{fig:remove-fraction-softlabel}
\end{figure}

In Fig.~\ref{fig:remove-fraction-softlabel}(A) we observe the effect of removing parts of the logits. Already removing a single entry is critical when the normal student would otherwise have learned the teacher. Removing more deteriorates performance in accuracy quickly, indicating that accessing the complete soft label is important to recover held-out memorized items.

In Fig.~\ref{fig:remove-fraction-softlabel}(B) we observe that  removing the class $c$ from the soft labels is detrimental to the accuracy on that class in the test set ($\accZeroStudent$) but is maintained almost at a normal level for other classes. The average performance on the other hand is affected only little.
In contrast, when we remove the class $c$ completely from the training set, but leave it intact in other classes' soft labels, the held-out sample accuracy remains at a high level for class $c$, as well as the others.
This further emphasizes that a lot of information is contained in the soft labels, and that especially the relational information to the class can help a lot.

\newpage
\subsubsection{Varying the hidden layer size of teachers and students}

\begin{figure}[h!]
    \centering
    \includegraphics[width=0.5\linewidth]{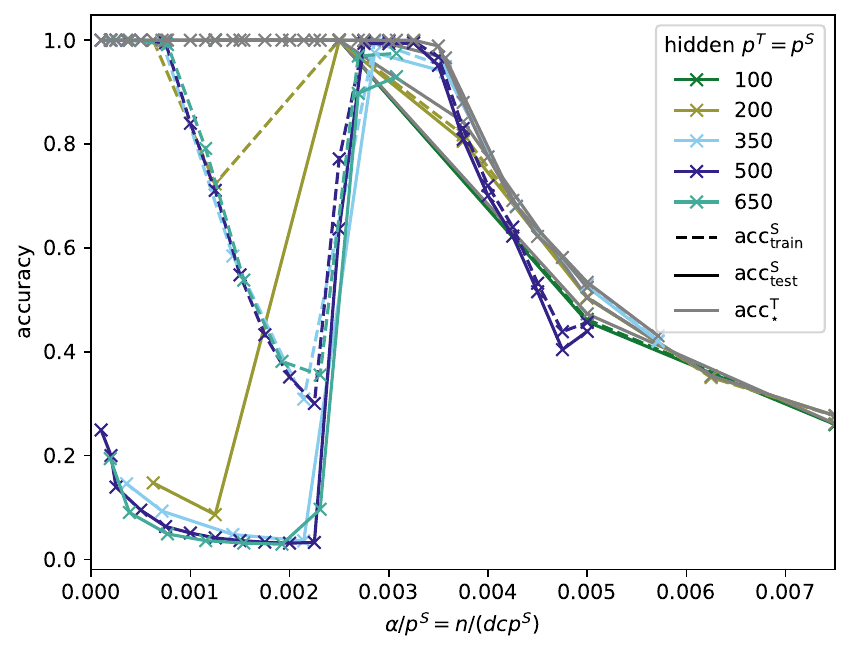}
    \caption{For ReLU MLPs we vary the teacher and students hidden layer sizes $p^T = p^S$ jointly. We keep $\rho=0.45$, $c=100$, $d=1000$, and vary $n \in d \cdot \{12.5,  25,  50,  75, 100, 125, 150, 17, 200\}$, with hidden layer sizes as in the legend. Experiments here are repeated once.}
    \label{fig:change-hidden-together}
\end{figure}

\begin{figure}[h!]
    \centering
    \includegraphics[width=0.5\linewidth]{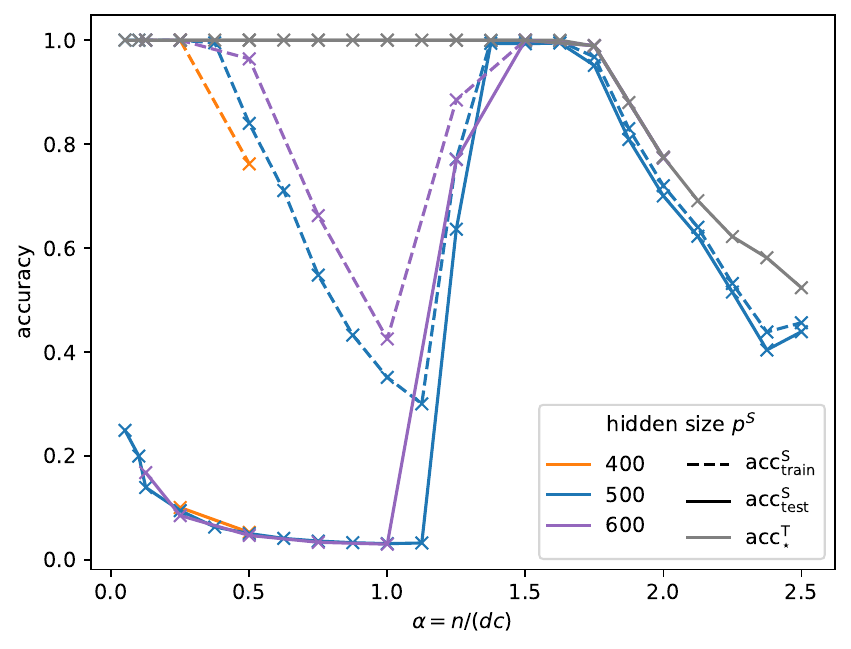}
    \caption{For ReLU MLPs we keep the teacher hidden layer sizes $p^T=500$ and vary the studens hidden layer size $p^S$. We keep $\rho=0.45$, $c=100$, $d=1000$, and vary $n$ with hidden layer sizes as in the legend. Experiments here are repeated once.}
    \label{fig:change-hidden-student}
\end{figure}

In the following, we vary the sizes of the hidden layers in the teacher and student single layer ReLU MLPs, calling them $p^T$ and $p^S$ respectively.
We first keep $\rho=0.45$ fixed and keep the temperature $\temp=20$.
In Fig.~\ref{fig:change-hidden-together} and \ref{fig:change-hidden-student}, we vary $p^T$ and $p^S$ in isolation or together respectively.

In running our experiments for Fig.~\ref{fig:change-hidden-together} we keep $d$ constant and vary $n$ and $p^T=p^S$, but instead of plotting the resulting accuracies over $\alpha=n/(dc)$ as in the main we plot them over $\alpha / p^S$. This incorporates the hidden layer size $p$ in the denominator that represents the number of parameters of the model, and indeed the curves fall together quite accurately.

In Fig.~\ref{fig:change-hidden-together} we repeat the same experiment, but now keeping $p^T = 500$ and varying the student $p^S$.
As we can see, this improves $\accTrainStudent$ for e.g. $\rho=0.7$. This is expected as the student with more neurons has a higher capacity to learn the soft labels, which is the phase for the given $\rho$.

\begin{figure}[h!]
    \centering
    \includegraphics[width=0.5\linewidth]{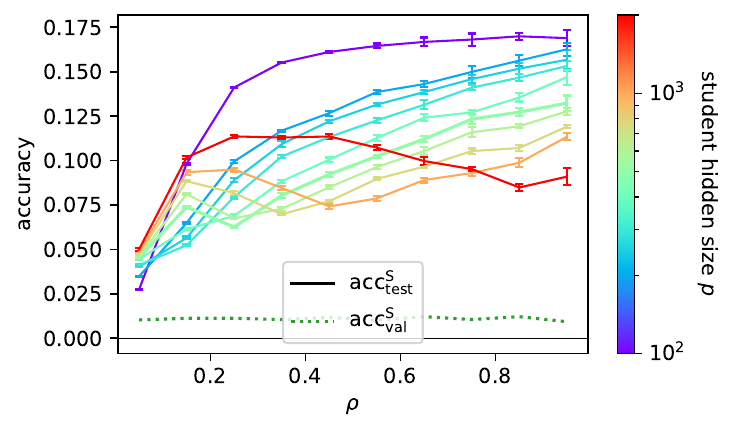}
    \caption{The teacher hidden layer size for the ReLU MLP is kept at $p^T=500$, while we vary the data size fraction $\rho$ and the student hidden layer size $p^S \in \{100, 200, 250, 300, 400, 500, 600, 750, 1000, 2000\}$. We report the test accuracy and $\accValStudent$ over the student with $p^S=500$. Experiments are repeated 5 times and the standard error on the mean is reported.}
    \label{fig:memorization-vary-capacity}
\end{figure}

\newpage
In Fig.~\ref{fig:memorization-vary-capacity} we zoom into the phase where the student is not matching the teacher but still finding a non-trivial $\accTestStudent > \accValStudent$ on the held-out memorized teacher data.
While we keep the teacher size fixed at $p^T=500$ we observe that lowering the capacity of the student is optimal for all sizes of the dataset. On the other hand, for larger $p^S$ the accuracy decreases but not in a linear way - e.g. for $\rho=0.3$ students with $p^S=250$ are as good as students with $p^S=1000$.

\newpage
\subsubsection{Varying the number of classes}
\begin{figure}[h!]
    \centering
    \includegraphics[width=0.5\linewidth]{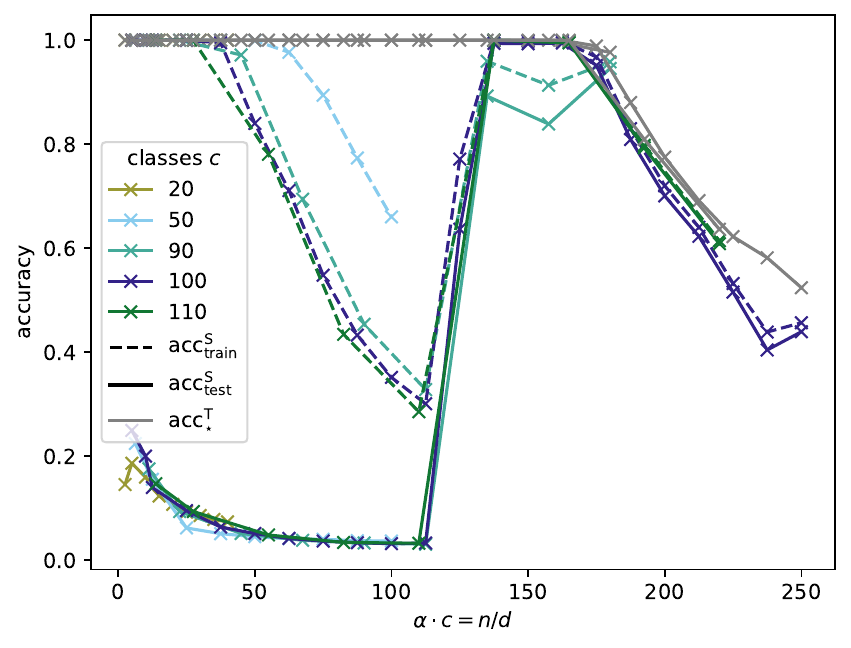}
    \caption{For ReLU MLPs with $p=500$ for both teacher and student, we vary the number of classes $c \in \{20,50,90,100,110\}$ and show the student's accuracy on $\Dstudenttrain, \Dstudenttest$ in a single run, and compare with the teacher accuracy.}
    \label{fig:mlp-classes}
\end{figure}

Finally, we examine the effect of the number of classes in Fig.~\ref{fig:mlp-classes}. When we scale the $x$-axis as $n/d$, the behavior we described in the main for the different classes remains similar.

\section{Computational Resources}

All experiments can be run both on a CPU or GPU - for multinomial logistic regression a CPU may be faster than a GPU.\\
The most computational intensive were the phase diagrams Fig. 4a) with 13 compute days, and Fig. 6a) with roughly 10 compute days on a GPU. In both cases, we ran the full pipeline parallelizing experiments on a single machine with an NVIDIA RTX A5000.\\
Since many of the experiments are run for different seeds to obtain error bars, running the experiments once is roughly a factor 5 faster than running all. 

\end{document}